\documentclass[11pt]{article}

\usepackage[final]{acl}

\usepackage{times}
\usepackage{latexsym}

\usepackage[T1]{fontenc}

\usepackage[utf8]{inputenc}

\usepackage{microtype}

\usepackage{inconsolata}

\usepackage{graphicx}

\usepackage{microtype}
\usepackage{graphicx}
\usepackage{subcaption}
\usepackage{booktabs} %
\usepackage{multirow}
\usepackage{amsmath}
\usepackage{amssymb}
\usepackage{mathtools}
\usepackage{amsthm}
\usepackage{tabularray}
\usepackage{algorithm}
\usepackage{algpseudocode}
\usepackage[capitalize,noabbrev]{cleveref}

\newif\ifshowrev
\showrevfalse
\DeclareRobustCommand{\rev}[1]{\ifshowrev{\color{blue}#1}\else#1\fi}
\newcommand{\revcolor}{\ifshowrev\color{blue}\fi}  %
\AtBeginDocument{\pdfstringdefDisableCommands{\def\rev#1{#1}}}  %

\theoremstyle{plain}

\theoremstyle{definition}

\theoremstyle{remark}

\title{Dataset-Level Metrics \rev{Mask} Non-Determinism: A Fine-Grained Non-Determinism Evaluation in Diffusion Language Models}

\author{
  \textbf{Zhengyu Fang\textsuperscript{1}},
  \textbf{Zhimeng Jiang\textsuperscript{2}},
  \textbf{Huiyuan Chen\textsuperscript{1}},
  \textbf{Xiaoge Zhang\textsuperscript{1}},
\\
  \textbf{Tianyi Li\textsuperscript{1}},
  \textbf{Kaiyu Tang\textsuperscript{1}},
  \textbf{Xiao Li\textsuperscript{1}},
  \textbf{Jing Li\textsuperscript{1}}
\\
  \textsuperscript{1}Case Western Reserve University \quad
  \textsuperscript{2}Texas A\&M University
\\
  \texttt{\{zxf177,hxc501,xxz705,txl805,kxt439,xxl1015,jingli\}@case.edu}
\\
  \texttt{zhimengj@tamu.edu}
}

\begin{document}
\maketitle
\begin{abstract}
Diffusion language models (DLMs) have emerged as a promising paradigm for large language models (LLMs), yet the non-deterministic behavior of DLMs remains poorly understood. \rev{Existing} non-determinism evaluations for LLMs predominantly rely on dataset-level metrics under fixed inference configurations, providing limited insight into how model behavior varies across runs and evaluation conditions. In this work, we show that dataset-level metrics systematically \textbf{\rev{mask} non-determinism} in \rev{LLaDA-family DLMs} by \textit{aggregating sample-level prediction quality} across different runs\rev{: configurations with similar aggregate performance can behave substantially differently on individual inputs. Distinguishing strict (fixed-protocol) non-determinism from system-induced and decoding-induced non-determinism,} we conduct a fine-grained evaluation based on \textbf{sample-level prediction differences} across \rev{model-related factors (guidance scale, diffusion steps, Monte Carlo sampling), system-related factors (batch size, hardware, numerical precision), and fixed-protocol repetitions}. Our analysis \rev{of two LLaDA-family checkpoints} reveals that \rev{configuration-induced} non-determinism is pervasive and structured, with code generation exhibiting markedly higher sensitivity to \rev{configuration changes} than question answering. \rev{To characterize how non-determinism is distributed across factors,} we introduce Factor Variance Attribution (FVA), \rev{an ANOVA-based diagnostic that contrasts between-factor and within-factor variation on the evaluated configuration grid}. Our findings highlight the need for fine-grained, factor-aware evaluation to enable reliable non-determinism assessment of \rev{DLMs}.
\end{abstract}

\section{Introduction}

Non-determinism has become an increasingly important concern in the evaluation of large language models~\cite{yuan2025understanding,he2025nondeterminism}: repeated evaluations under the same conditions can yield inconsistent predictions, raising questions about reliability, robustness, and reproducibility. While prior work has examined non-determinism primarily through variability in dataset-level performance metrics, such analyses offer only a coarse view of model non-determinism behavior and provide limited insight into non-determinism measurement. \rev{We use non-determinism as an umbrella term and distinguish three sources: (i) \textit{strict non-determinism}, i.e., fixed-protocol run-to-run stochasticity, where the model, decoding configuration, system, and seed policy are held fixed; (ii) \textit{system-induced non-determinism}, where hardware, numerical precision, or execution order changes while decoding is fixed; and (iii) \textit{decoding-induced non-determinism}, where choices such as guidance scale, diffusion steps, or the Monte Carlo budget are deliberately changed. We refer to (ii) and (iii) jointly as \textit{configuration-induced non-determinism}.}

This challenge is particularly acute for diffusion language models (DLMs)~\cite{austin2021structured,nie2025large,zhu2025llada,ye2025dream}. Unlike autoregressive decoding, diffusion-based generation is inherently stochastic and proceeds through multiple inference steps, introducing additional degrees of freedom at inference time, including sampling strategies, guidance scales, diffusion steps, and numerical execution details~\cite{li2025survey}. As a result, DLMs can exhibit pronounced \rev{configuration-induced} non-determinism even when evaluated under seemingly minor configuration changes. Crucially, different inference configurations can achieve nearly identical dataset-level metrics, such as accuracy or pass@k~\cite{chen2021evaluating,liang2022holistic,touvron2023llama}, while producing substantially different predictions for the same individual inputs.

We argue that this phenomenon exposes a fundamental limitation of existing evaluation practices. Dataset-level metrics aggregate sample-level prediction quality across inputs in different runs, thereby \rev{masking} non-determinism by reducing sensitivity to input-conditional variability. When correctness systematically flips for individual samples across configurations (e.g., one \rev{wrong-to-right} flip and one right-to-wrong), these changes are averaged out, obscuring which inputs are affected, how error modes differ, and whether observed variability arises from specific inference factors or from instability within fixed settings~\cite{sculley2015hidden}. Consequently, evaluation based solely on aggregate quality metrics can give a misleading impression of stability and reproducibility, particularly for diffusion-based generation.

In this work, we propose a fine-grained, cross-factor evaluation framework for characterizing non-determinism in diffusion language models beyond dataset-level metrics. \textbf{First}, we adopt a sample-level evaluation perspective that explicitly measures how prediction outcomes for the same input vary across inference-time configurations, enabling the identification of correctness flips that are invisible to dataset-level metrics. \textbf{Second}, we systematically study non-determinism induced by both \textit{model-related factors}, which directly affect diffusion dynamics \rev{(decoding-induced)}, and \textit{system-related factors}, which arise from execution, hardware, and numerical choices \rev{(system-induced)}, without any performance-driven tuning or hyperparameter optimization. \rev{We complement these one-factor-at-a-time sweeps with fixed-protocol repetitions that directly assess strict non-determinism.} \textbf{Third}, we introduce \textit{Factor Variance Attribution (FVA)}, \rev{which combines standard one-way ANOVA mean squares, rather than a new estimator, into a single contrast between} non-determinism arising from the choice of evaluation factors \rev{and} sensitivity to specific settings within each factor, enabling \rev{an} analysis of task-dependent evaluation non-determinism beyond dataset-level metrics. \rev{Because each factor is varied around a reference configuration, FVA summarizes reference-conditional first-order sensitivity on the evaluated grid; it is a descriptive diagnostic rather than a causal or population-level attribution.}

Our empirical analysis \rev{of two checkpoints from the LLaDA model family} reveals that \rev{configuration-induced} non-determinism is pervasive and highly structured. In particular, code generation tasks exhibit substantially greater sensitivity to \rev{configuration changes} than question answering, and the dominant sources of variability differ across tasks and evaluation regimes. \rev{Under a fixed protocol with identical random streams, repeated runs were exactly reproducible, whereas changing only the evaluation seed altered PIQA predictions but not HumanEval outputs.} Together, these findings demonstrate that dataset-level metrics alone provide an incomplete picture of non-determinism and highlight the need for fine-grained, cross-factor evaluation to support reliable non-determinism assessment of diffusion language models.

\begin{table*}[t]
\centering

\resizebox{1.0\textwidth}{!}{%
\begin{tabular}{p{7.2cm} p{7.2cm} c c c p{1.8cm}}
\toprule
Question & Answer Options
& INT8 & BF16 & FP16 & Ground Truth \\
\midrule

In regions where prevailing winds blow from the oceans to the shore, which of these is most likely to occur?
&
A: heavier rainfall \newline
B: dryer conditions \newline
C: cold temperatures \newline
D: frequent hurricanes
&
D & C & C & A \\

Matt is a tall, eleven-year-old boy. He has a scar on his right cheek. He is intelligent and an excellent drummer.
Which of his traits did he most likely inherit?
&
A: his height \newline
B: his scar on his right cheek \newline
C: his intelligence \newline
D: his ability to play the drums
&
B & C & C & A \\

Two bicyclists were riding their bikes at a rate of 20 kilometers per hour (km/h).
During the next half hour, they observed that they had increased their speed to 26 km/h.
What was their average acceleration?
&
A: 0.2 km/h$^2$ \newline
B: 3.0 km/h$^2$ \newline
C: 6.0 km/h$^2$ \newline
D: 12.0 km/h$^2$
&
A & C & C & D \\

\bottomrule
\end{tabular}
}
\caption{
Examples of sample-level prediction disagreements induced by numerical precision (LLaDA).
All configurations contribute identically at the dataset level,
yet produce different answers for the same inputs,
exposing distinct error modes that are obscured by aggregate metrics.
Additional qualitative examples are provided in Appendix~\ref{sec:case-studies}.
}
\label{tab:sample-level-disagreement-with-gt}
\end{table*}

\begin{figure*}[t]
    \centering
    \begin{subfigure}[t]{0.33\textwidth}
        \centering
        \includegraphics[width=\linewidth]{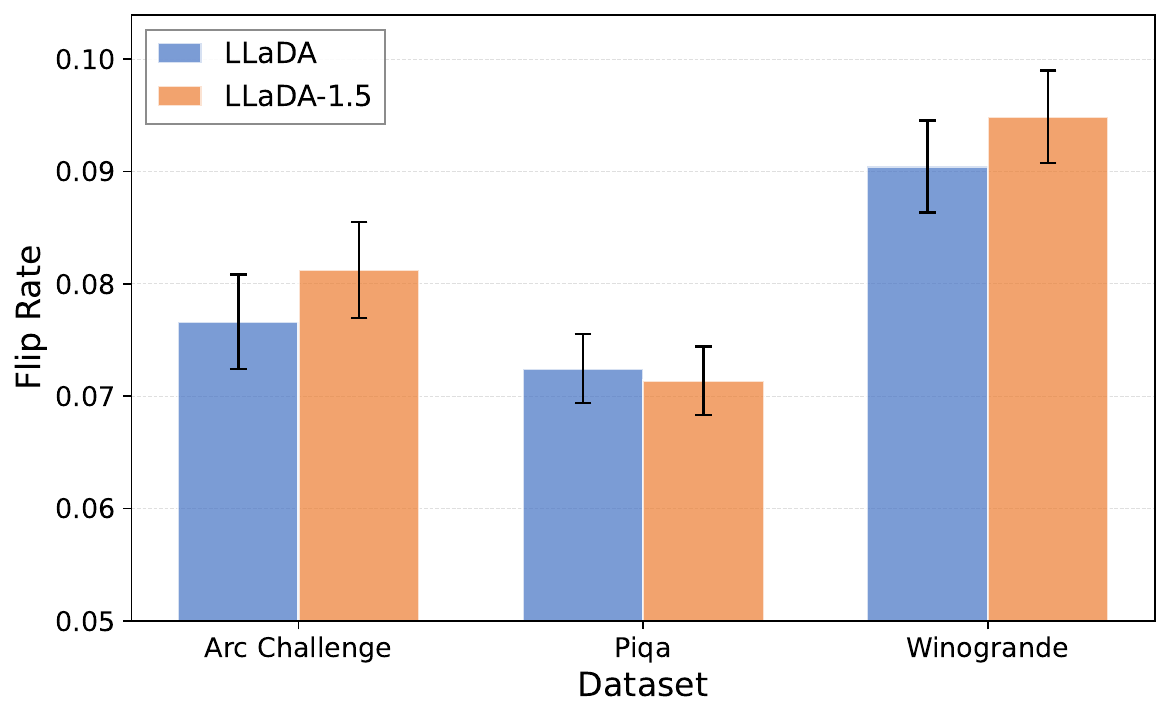}
        \caption{Batch size.}
        \label{fig:prediction-flip-rate-batch}
    \end{subfigure}\hfill
    \begin{subfigure}[t]{0.33\textwidth}
        \centering
        \includegraphics[width=\linewidth]{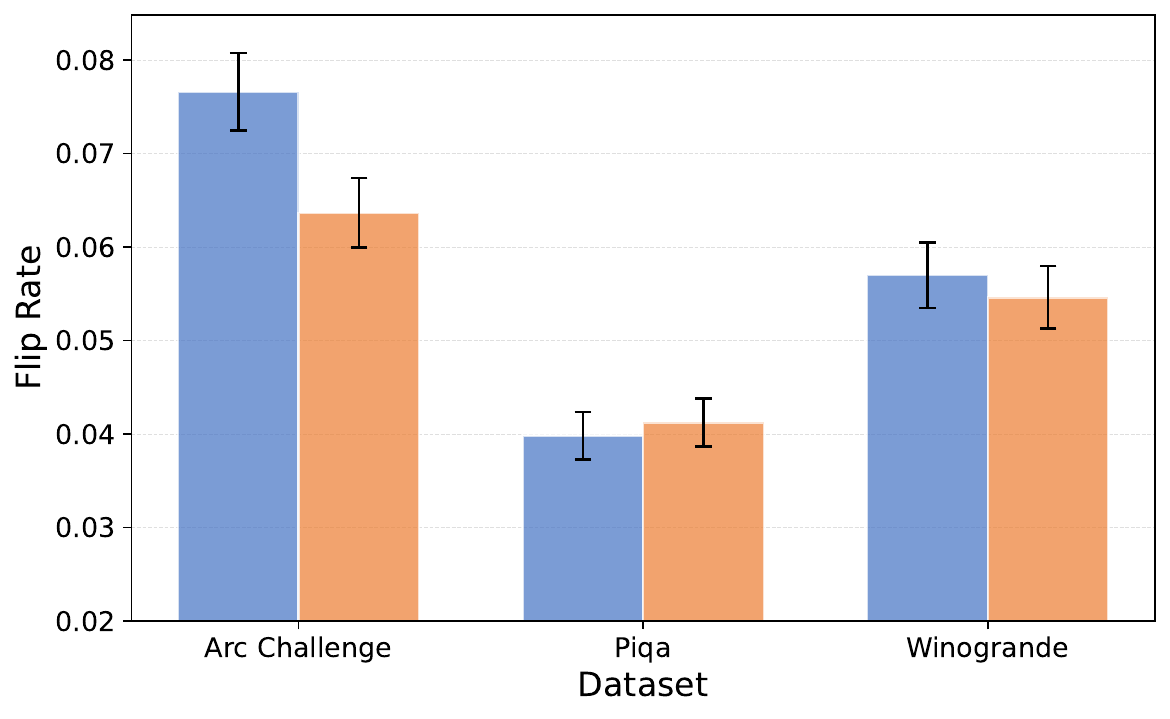}
        \caption{CFG scale.}
        \label{fig:prediction-flip-rate-cfg}
    \end{subfigure}\hfill
    \begin{subfigure}[t]{0.33\textwidth}
        \centering
        \includegraphics[width=\linewidth]{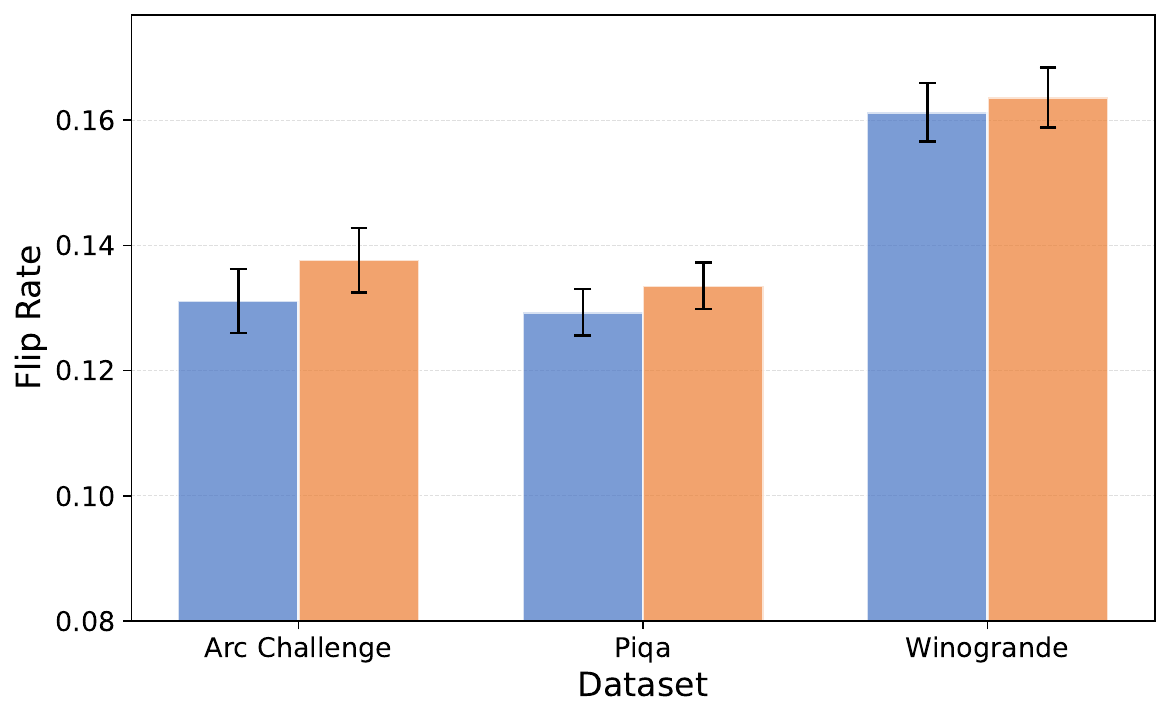}
        \caption{Monte Carlo samples.}
        \label{fig:prediction-flip-rate-mc}
    \end{subfigure}

    \caption{
Sample-level prediction flip rates across inference-time configurations.
Each panel varies one inference factor while holding all other factors fixed:
\textbf{(a)} batch size,
\textbf{(b)} classifier-free guidance (CFG) scale,
and \textbf{(c)} number of Monte Carlo (MC) samples.
For a given sample $s$, the prediction flip rate is defined as
$1 - \max_{a} \lvert \{ c \in \mathcal{C} : \hat{y}_{s,c} = a \} \rvert / \lvert \mathcal{C} \rvert$,
where $\hat{y}_{s,c}$ denotes the predicted label for sample $s$ under configuration $c$,
$a$ ranges over all possible predictions,
and $\mathcal{C}$ is the set of configurations obtained by varying the selected factor across its settings, with all other factors held fixed.
Reported values are averages of the per-sample flip rates over the dataset,
with error bars indicating the standard error across samples.
Larger flip rates indicate stronger configuration-induced non-determinism at the sample level,
even when dataset-level accuracy remains nearly unchanged.
}
    \label{fig:prediction-flip-rate}
    \vspace{-10pt}
\end{figure*}

\section{Related Work}

\subsection{Diffusion Language Models and Evaluation}

Diffusion language models (DLMs) have recently emerged as an alternative to autoregressive generation, extending diffusion and denoising paradigms to language modeling and a broad range of generative tasks~\cite{li2025survey}. This line of work now spans continuous-space models~\cite{li2022diffusion,gong2022diffuseq,dieleman2022continuous,yuan2022seqdiffuseq,han2023ssd,lovelace2023latent,mahabadi2024tess,tae2025tess}, discrete masked diffusion models~\cite{austin2021structured,he2023diffusionbert,sahoo2024simple,gat2024discrete,ye2025dream,nie2025large,liu2025longllada,gong2025diffucoder,labs2025mercury,zhu2025llada,cheng2025sdar,liu2025tidar,liu2025sequential}, and multimodal diffusion language models~\cite{you2025llada,yang2025mmada,li2025lavida,yu2025dimple,xin2025lumina}. Recent large-scale DLMs have demonstrated competitive performance on tasks such as question answering, reasoning, and code generation, while offering distinct properties such as parallel decoding and flexible inference-time control.

Despite these differences in generation mechanisms, evaluation practices for diffusion language models largely follow those used for autoregressive language models. Performance is typically reported under a fixed inference-time configuration using dataset-level metrics such as accuracy, exact match, or pass@k~\cite{chen2021evaluating,liu2023your,touvron2023llama,chang2024survey,yuan2025science}. While effective for benchmarking overall capability, these metrics emphasize aggregate performance and implicitly treat inference-time choices, such as diffusion steps, guidance strength, sampling strategies, or numerical precision, as incidental to the evaluation, rather than objects of analysis.

\subsection{Reproducibility and Stability in Generative Models}

Reproducibility and stability have long been recognized as challenges in machine learning systems, particularly for models involving stochastic training or inference~\cite{sculley2015hidden}. In the context of large language models, prior work has examined sources of \rev{non-determinism} arising from random seeds, sampling procedures, numerical precision, and hardware execution, showing that such factors can lead to nontrivial variability in model outputs~\cite{yu2023benchmarking,yuan2025understanding,bui2025assessing,klishevich2025measuring,ouyang2025empirical,wang2025assessing,zeng2025analyst}.
Beyond run-to-run variance, several studies have noted that models with similar aggregate metrics can exhibit inconsistent behavior on individual inputs or generations, raising concerns about robustness and reproducibility~\cite{ribeiro2020beyond,mehrabi2021survey}. These works highlight that average performance alone may be insufficient to characterize model reliability, especially during stochastic inference.

However, most prior work focuses on output diversity, run-to-run variability, or distributional effects of randomness~\cite{zhang2025beyond}, rather than \emph{correctness stability} at the sample level under different inference-time configurations. Our work complements this literature by providing a sample-level analysis of evaluation non-determinism and by \rev{contrasting variability across} classes of inference-time factors, bridging reproducibility concerns with configuration-aware evaluation.

\section{Preliminary Analysis and Fine-Grained Evaluation Paradigm}
\subsection{Preliminary Analysis}

\begin{figure*}[t]
    \centering
    \includegraphics[width=0.75\textwidth]{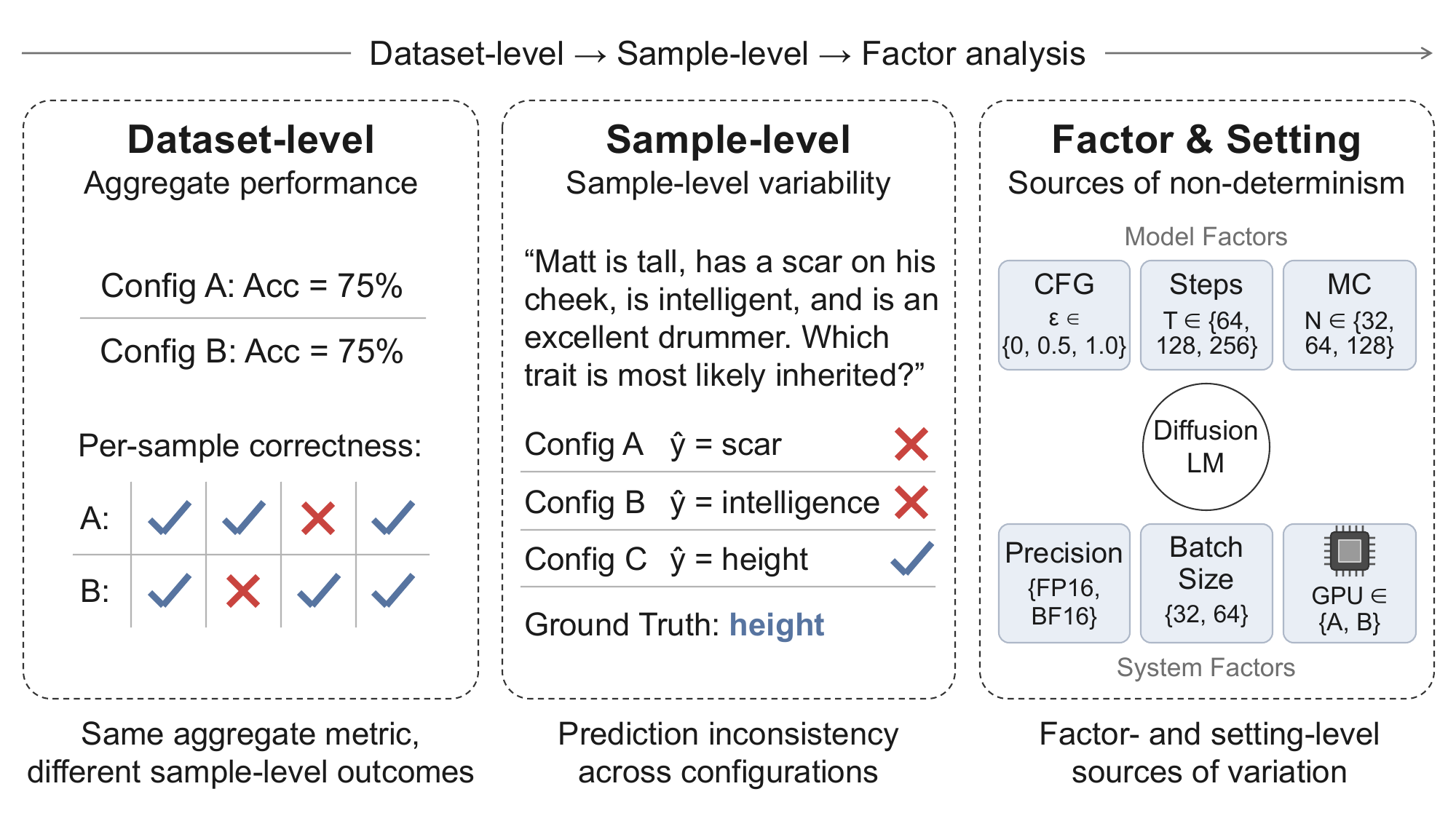}
    \caption{
    This hierarchy underlies our sample-aware and factor-aware evaluation paradigm
and is formally instantiated by the definitions in \Cref{sec:evaluation-paradigm}.
    }
    \label{fig:evaluation-paradigm}
    \vspace{-10pt}
\end{figure*}
\paragraph{Dataset-Level Stability Can \rev{Mask} Substantial Non-Determinism.}

Standard evaluation of diffusion language models~\rev{\cite{li2025survey}} typically reports dataset-level output quality under a fixed inference-time configuration, implicitly treating the aggregate score as representative of model behavior. However, this practice can obscure substantial variability at the level of individual samples\footnote{Compared to autoregressive models, non-determinism in DLMs has been much less studied. We therefore start from DLMs to highlight this issue and encourage more attention from the community.}.

\Cref{fig:prediction-flip-rate} provides a first empirical illustration of this phenomenon.
Across multiple benchmarks, we observe nontrivial sample-level prediction flip rates when varying inference-time factors,
including batch size, classifier-free guidance (CFG) scale, and the number of Monte Carlo (MC) samples,
even though dataset-level accuracy remains largely unchanged.
In many cases, the same input yields different predicted answers across configurations,
despite all configurations being \rev{consistently incorrect}.
This reveals a form of configuration-induced non-determinism at the level of predicted outputs that is invisible to correctness-based aggregate metrics.

Complementary qualitative examples in \Cref{tab:sample-level-disagreement-with-gt,tab:sample-level-disagreement-mc} further illustrate this behavior.
These examples span both system-related factors, such as numerical precision, and model-related factors, such as Monte Carlo sampling.
Although the corresponding configurations are indistinguishable under dataset-level accuracy, they often correspond to different incorrect answers, indicating distinct failure modes for the same input.
By collapsing all outcomes into a single scalar, dataset-level metrics discard information about which samples change across configurations and how their errors differ~\cite{ribeiro2020beyond}.
The consistency of this phenomenon across factors suggests that such non-determinism is not an isolated artifact, but a recurring property of \rev{evaluation for the LLaDA-family checkpoints we study}.

\paragraph{Towards Factor-Aware and Sample-Aware Evaluation.}

The preceding observations indicate that evaluation variability in \rev{the evaluated} diffusion language models cannot be adequately characterized by aggregate metrics alone~\cite{ribeiro2020beyond,dror2018hitchhiker}.
When different inference-time configurations yield identical dataset-level performance yet diverge in their sample-level behavior and error modes, treating configuration choices as mere implementation details becomes insufficient.
Instead, configuration choices form an integral part of the evaluation protocol and must be analyzed explicitly.

This motivates an evaluation perspective that distinguishes between two conceptually distinct sources of variability:
\emph{factor-level effects}, arising from which dimension of the evaluation protocol is varied (e.g., sampling strategy or numerical precision),
and \emph{setting-level effects}, arising from the specific value chosen within a factor (e.g., the number of Monte Carlo samples or the guidance scale).
Disentangling these sources is essential for interpreting reported results, assessing robustness, and understanding why seemingly similar evaluations can lead to qualitatively different conclusions~\cite{searle2009variance,koo2016guideline}.
Accordingly, we adopt a sample-aware and factor-aware evaluation framework that explicitly decomposes evaluation variability along these dimensions.

\begin{table*}[t]
\scriptsize
\centering
\resizebox{\textwidth}{!}{%
\begin{tabular}{llcccccc}
\toprule
Dataset & Factor & \rev{$\bar a$} & \rev{$s_a$} & \rev{$\mathrm{Std}_x$} & \rev{$V_{\mathrm{item}}$ [95\% CI]} & \rev{$D_{\mathrm{corr}}$ [95\% CI]} & \rev{$D_{\mathrm{out}}$ [95\% CI]} \\
\midrule
\multicolumn{8}{c}{\textbf{LLaDA}} \\
\midrule
\multirow{5}{*}{PIQA} & Precision & 0.7347 & 0.0006 & 0.4381 & \rev{0.0031\,[0.0019,0.0044]} & \rev{0.0094\,[0.0058,0.0131]} & \rev{0.0094\,[0.0058,0.0131]} \\
 & Batch Size of MC & 0.7383 & 0.0152 & 0.3790 & \rev{0.0497\,[0.0458,0.0540]} & \rev{0.1242\,[0.1145,0.1349]} & \rev{0.1242\,[0.1145,0.1349]} \\
 & CFG & 0.7274 & 0.0068 & 0.4148 & \rev{0.0263\,[0.0232,0.0296]} & \rev{0.0657\,[0.0580,0.0739]} & \rev{0.0657\,[0.0580,0.0739]} \\
 & \# of MC & 0.7087 & 0.0398 & 0.3454 & \rev{0.0872\,[0.0829,0.0915]} & \rev{0.2093\,[0.1989,0.2195]} & \rev{0.2093\,[0.1989,0.2195]} \\
 & GPUs & 0.7342 & 0.0012 & 0.4405 & \rev{0.0012\,[0.0005,0.0020]} & \rev{0.0049\,[0.0022,0.0082]} & \rev{0.0049\,[0.0022,0.0082]} \\
\addlinespace[2pt]
\multirow{5}{*}{WinoGrande} & Precision & 0.7340 & 0.0014 & 0.4382 & \rev{0.0033\,[0.0019,0.0049]} & \rev{0.0100\,[0.0058,0.0147]} & \rev{0.0100\,[0.0058,0.0147]} \\
 & Batch Size of MC & 0.7291 & 0.0148 & 0.3694 & \rev{0.0611\,[0.0562,0.0664]} & \rev{0.1528\,[0.1405,0.1659]} & \rev{0.1528\,[0.1405,0.1659]} \\
 & CFG & 0.7263 & 0.0074 & 0.4011 & \rev{0.0380\,[0.0337,0.0421]} & \rev{0.0950\,[0.0843,0.1053]} & \rev{0.0950\,[0.0843,0.1053]} \\
 & \# of MC & 0.7027 & 0.0455 & 0.3190 & \rev{0.1073\,[0.1016,0.1131]} & \rev{0.2574\,[0.2439,0.2714]} & \rev{0.2574\,[0.2439,0.2714]} \\
 & GPUs & 0.7348 & 0.0000 & 0.4398 & \rev{0.0016\,[0.0006,0.0028]} & \rev{0.0063\,[0.0024,0.0110]} & \rev{0.0063\,[0.0024,0.0110]} \\
\addlinespace[2pt]
\multirow{5}{*}{ARC-Challenge} & Precision & 0.4480 & 0.0023 & 0.4931 & \rev{0.0044\,[0.0027,0.0064]} & \rev{0.0131\,[0.0080,0.0193]} & \rev{0.0233\,[0.0159,0.0307]} \\
 & Batch Size of MC & 0.4370 & 0.0082 & 0.4644 & \rev{0.0305\,[0.0266,0.0345]} & \rev{0.0763\,[0.0665,0.0863]} & \rev{0.1286\,[0.1161,0.1412]} \\
 & CFG & 0.4369 & 0.0072 & 0.4625 & \rev{0.0323\,[0.0281,0.0366]} & \rev{0.0807\,[0.0703,0.0915]} & \rev{0.1282\,[0.1147,0.1409]} \\
 & \# of MC & 0.4242 & 0.0233 & 0.4349 & \rev{0.0553\,[0.0504,0.0604]} & \rev{0.1327\,[0.1210,0.1449]} & \rev{0.2109\,[0.1958,0.2256]} \\
 & GPUs & 0.4288 & 0.0018 & 0.4927 & \rev{0.0023\,[0.0009,0.0038]} & \rev{0.0094\,[0.0034,0.0154]} & \rev{0.0128\,[0.0068,0.0196]} \\
\midrule
\multicolumn{8}{c}{\textbf{LLaDA-1.5}} \\
\midrule
\multirow{5}{*}{PIQA} & Precision & 0.7486 & 0.0005 & 0.4315 & \rev{0.0021\,[0.0011,0.0030]} & \rev{0.0062\,[0.0033,0.0091]} & \rev{0.0062\,[0.0033,0.0091]} \\
 & Batch Size of MC & 0.7441 & 0.0182 & 0.3760 & \rev{0.0491\,[0.0455,0.0534]} & \rev{0.1229\,[0.1137,0.1334]} & \rev{0.1229\,[0.1137,0.1334]} \\
 & CFG & 0.7430 & 0.0045 & 0.4043 & \rev{0.0276\,[0.0244,0.0309]} & \rev{0.0689\,[0.0610,0.0773]} & \rev{0.0689\,[0.0610,0.0773]} \\
 & \# of MC & 0.7166 & 0.0423 & 0.3364 & \rev{0.0900\,[0.0854,0.0946]} & \rev{0.2160\,[0.2050,0.2270]} & \rev{0.2160\,[0.2050,0.2270]} \\
 & GPUs & 0.7489 & 0.0004 & 0.4314 & \rev{0.0020\,[0.0011,0.0031]} & \rev{0.0082\,[0.0044,0.0125]} & \rev{0.0082\,[0.0044,0.0125]} \\
\addlinespace[2pt]
\multirow{5}{*}{WinoGrande} & Precision & 0.7203 & 0.0030 & 0.4419 & \rev{0.0063\,[0.0044,0.0082]} & \rev{0.0189\,[0.0132,0.0247]} & \rev{0.0189\,[0.0132,0.0247]} \\
 & Batch Size of MC & 0.7149 & 0.0134 & 0.3734 & \rev{0.0645\,[0.0592,0.0696]} & \rev{0.1613\,[0.1479,0.1741]} & \rev{0.1613\,[0.1479,0.1741]} \\
 & CFG & 0.7059 & 0.0128 & 0.4127 & \rev{0.0374\,[0.0333,0.0417]} & \rev{0.0936\,[0.0833,0.1042]} & \rev{0.0936\,[0.0833,0.1042]} \\
 & \# of MC & 0.6885 & 0.0366 & 0.3261 & \rev{0.1082\,[0.1027,0.1137]} & \rev{0.2597\,[0.2466,0.2729]} & \rev{0.2597\,[0.2466,0.2729]} \\
 & GPUs & 0.7245 & 0.0011 & 0.4398 & \rev{0.0063\,[0.0041,0.0085]} & \rev{0.0253\,[0.0166,0.0339]} & \rev{0.0253\,[0.0166,0.0339]} \\
\addlinespace[2pt]
\multirow{5}{*}{ARC-Challenge} & Precision & 0.5267 & 0.0025 & 0.4963 & \rev{0.0032\,[0.0017,0.0049]} & \rev{0.0097\,[0.0051,0.0148]} & \rev{0.0165\,[0.0108,0.0228]} \\
 & Batch Size of MC & 0.5329 & 0.0068 & 0.4610 & \rev{0.0366\,[0.0319,0.0413]} & \rev{0.0915\,[0.0797,0.1032]} & \rev{0.1377\,[0.1250,0.1516]} \\
 & CFG & 0.5304 & 0.0099 & 0.4703 & \rev{0.0281\,[0.0238,0.0322]} & \rev{0.0701\,[0.0596,0.0806]} & \rev{0.1095\,[0.0972,0.1222]} \\
 & \# of MC & 0.5095 & 0.0220 & 0.4354 & \rev{0.0605\,[0.0553,0.0656]} & \rev{0.1452\,[0.1328,0.1574]} & \rev{0.2232\,[0.2080,0.2387]} \\
 & GPUs & 0.5418 & 0.0024 & 0.4972 & \rev{0.0013\,[0.0002,0.0023]} & \rev{0.0051\,[0.0009,0.0094]} & \rev{0.0060\,[0.0017,0.0102]} \\
\bottomrule
\end{tabular}
}
\caption{\rev{Matched-item evaluation results on question answering datasets for two LLaDA-family checkpoints. For each factor, all statistics are computed on the complete-case matched $N\times C$ grid (\Cref{sec:evaluation-paradigm}). $\bar a$ and $s_a$ are the mean and sample standard deviation of aggregate accuracy across the $C$ configurations. $\mathrm{Std}_x$ is the cross-item sample standard deviation of each item's configuration-averaged correctness; it reflects heterogeneity across items rather than within-item change across configurations. $V_{\mathrm{item}}$ is the mean within-item correctness variance, and $D_{\mathrm{corr}}$ and $D_{\mathrm{out}}$ are the mean pairwise correctness and predicted-label disagreement across configuration pairs. Brackets give 95\% percentile intervals from 1{,}000 unique-item bootstrap replicates.}}
\label{tab:qa-results-backbone}
\end{table*}

\begin{table*}[t]
\scriptsize
\centering
\resizebox{\textwidth}{!}{%
\begin{tabular}{llcccccc}
\toprule
Dataset & Factor & \rev{$\bar a$} & \rev{$s_a$} & \rev{$\mathrm{Std}_x$} & \rev{$V_{\mathrm{item}}$ [95\% CI]} & \rev{$D_{\mathrm{corr}}$ [95\% CI]} & \rev{$D_{\mathrm{out}}$ [95\% CI]} \\
\midrule
\multicolumn{8}{c}{\textbf{LLaDA}} \\
\midrule
\multirow{4}{*}{HumanEval} & Precision & 0.3110 & 0.0061 & 0.4417 & \rev{0.0203\,[0.0108,0.0298]} & \rev{0.0610\,[0.0325,0.0894]} & \rev{0.4817\,[0.4187,0.5386]} \\
 & CFG & 0.2659 & 0.0432 & 0.3490 & \rev{0.0741\,[0.0600,0.0888]} & \rev{0.1854\,[0.1500,0.2220]} & \rev{0.9311\,[0.9018,0.9561]} \\
 & Steps & 0.1537 & 0.1258 & 0.2502 & \rev{0.0678\,[0.0532,0.0839]} & \rev{0.1695\,[0.1329,0.2098]} & \rev{0.9220\,[0.8915,0.9482]} \\
 & GPUs & 0.3323 & 0.0216 & 0.4643 & \rev{0.0076\,[0.0015,0.0152]} & \rev{0.0305\,[0.0061,0.0610]} & \rev{0.3354\,[0.2622,0.4024]} \\
\addlinespace[2pt]
\multirow{4}{*}{MBPP} & Precision & 0.4007 & 0.0083 & 0.4649 & \rev{0.0244\,[0.0182,0.0307]} & \rev{0.0733\,[0.0547,0.0920]} & \rev{0.4793\,[0.4467,0.5160]} \\
 & CFG & 0.2864 & 0.0776 & 0.3358 & \rev{0.0918\,[0.0834,0.1000]} & \rev{0.2296\,[0.2084,0.2500]} & \rev{0.9878\,[0.9822,0.9926]} \\
 & Steps & 0.1908 & 0.1534 & 0.2592 & \rev{0.0874\,[0.0781,0.0965]} & \rev{0.2184\,[0.1952,0.2412]} & \rev{0.9388\,[0.9288,0.9486]} \\
 & GPUs & 0.3990 & 0.0014 & 0.4783 & \rev{0.0115\,[0.0070,0.0160]} & \rev{0.0460\,[0.0280,0.0640]} & \rev{0.3380\,[0.2980,0.3760]} \\
\midrule
\multicolumn{8}{c}{\textbf{LLaDA-1.5}} \\
\midrule
\multirow{4}{*}{HumanEval} & Precision & 0.4106 & 0.0196 & 0.4621 & \rev{0.0298\,[0.0190,0.0420]} & \rev{0.0894\,[0.0569,0.1260]} & \rev{0.5203\,[0.4553,0.5772]} \\
 & CFG & 0.2646 & 0.1118 & 0.3396 & \rev{0.0800\,[0.0649,0.0956]} & \rev{0.2000\,[0.1621,0.2391]} & \rev{0.9189\,[0.8896,0.9452]} \\
 & Steps & 0.2454 & 0.1418 & 0.3114 & \rev{0.0888\,[0.0724,0.1056]} & \rev{0.2368\,[0.1931,0.2815]} & \rev{0.8984\,[0.8587,0.9350]} \\
 & GPUs & 0.3780 & 0.0259 & 0.4605 & \rev{0.0244\,[0.0137,0.0366]} & \rev{0.0976\,[0.0547,0.1463]} & \rev{0.5488\,[0.4756,0.6280]} \\
\addlinespace[2pt]
\multirow{4}{*}{MBPP} & Precision & 0.3860 & 0.0087 & 0.4701 & \rev{0.0164\,[0.0116,0.0218]} & \rev{0.0493\,[0.0347,0.0653]} & \rev{0.3427\,[0.3106,0.3787]} \\
 & CFG & 0.3052 & 0.0813 & 0.3609 & \rev{0.0821\,[0.0736,0.0914]} & \rev{0.2052\,[0.1840,0.2284]} & \rev{0.9630\,[0.9552,0.9702]} \\
 & Steps & 0.2630 & 0.1171 & 0.3458 & \rev{0.0745\,[0.0646,0.0839]} & \rev{0.1987\,[0.1723,0.2237]} & \rev{0.8880\,[0.8683,0.9063]} \\
 & GPUs & 0.3960 & 0.0226 & 0.4771 & \rev{0.0120\,[0.0075,0.0170]} & \rev{0.0480\,[0.0300,0.0680]} & \rev{0.3940\,[0.3520,0.4340]} \\
\bottomrule
\end{tabular}
}
\caption{\rev{Matched-item evaluation results on code generation datasets for two LLaDA-family checkpoints. Columns follow \Cref{tab:qa-results-backbone}, with pass@1 in place of accuracy. Here $D_{\mathrm{out}}$ compares the exact filtered candidate strings evaluated by the harness; it does not measure functional or semantic equivalence.}}
\label{tab:generation-results-backbone}
\end{table*}

\subsection{Fine-Grained Evaluation Paradigm}
\label{sec:evaluation-paradigm}

\paragraph{Sample-Level Evaluation View.}
We adopt a sample-level evaluation view, in which evaluation is defined at the level of individual inputs, as shown in \Cref{fig:evaluation-paradigm}.
Let $\mathcal{D} = \{(x_i, y_i)\}_{i=1}^N$ denote an evaluation dataset of $N$ samples, where $x_i$ is an input and $y_i$ is its ground-truth label.
Let $c \in \mathcal{C}$ denote an inference-time configuration, where $\mathcal{C}$ indexes all configurations under consideration.

For each sample $x_i$ and configuration $c$, we define a binary correctness indicator
$z_{i,c} \in \{0,1\}$, where $z_{i,c}=1$ if the model’s prediction under configuration $c$ is correct for sample $x_i$, and $z_{i,c}=0$ otherwise. \rev{For each factor grid, items are matched across configurations by document hash (falling back to dataset and document ID), and all statistics are computed on the complete-case intersection, so that the matched $N\times C$ matrix $Z=(z_{i,c})$ isolates variability induced by changes in inference-time configurations; all 46 model--dataset--factor grids retain every item. Items, rather than item--configuration pairs, are the sampling units for uncertainty quantification.}
The dataset-level accuracy under configuration $c$ is given by the empirical mean
$\frac{1}{N}\sum_{i=1}^N z_{i,c}$.
Standard evaluation focuses on this aggregate statistic.
Accordingly, dataset-level variability across configurations can be summarized by
$\mathrm{Var}_c\!\left( \frac{1}{N}\sum_{i=1}^N z_{i,c} \right)$,
which measures how overall accuracy changes as the configuration varies\rev{; we report its mean $\bar a$ and sample standard deviation $s_a$ (ddof$=1$) across configurations}.
In contrast, the sample-level view explicitly considers the variability of correctness for individual samples,
quantified by $\mathrm{Var}_c(z_{i,c})$, and its aggregation across samples.
Sample-level non-determinism arises when $\mathrm{Var}_c(z_{i,c})$ is large for a substantial fraction of samples, even when
$\mathrm{Var}_c\!\left( \frac{1}{N}\sum_{i=1}^N z_{i,c} \right)$ remains small.
This separation clarifies how evaluation can appear stable at the dataset level while masking substantial variability in per-sample behavior, including correctness flips and heterogeneous error modes across configurations.

\rev{We report two complementary reductions of $Z$. The direct within-item reduction averages each item's cross-configuration variance, $V_{\mathrm{item}}=\frac{1}{N}\sum_{i}\mathrm{Var}_c(z_{i,c})$, using the population convention (ddof$=0$) because the configurations form the complete evaluated grid. The cross-item reduction first averages over configurations, $\bar z_i=\frac{1}{C}\sum_c z_{i,c}$, and then takes the sample standard deviation across items ($\mathrm{Std}_x$, ddof$=1$); it incorporates each item's cross-configuration outcomes but measures heterogeneity across items. We additionally report the mean fraction of configuration pairs on which an item's correctness or prediction differs,
\begin{equation*}
\begin{aligned}
D_{\mathrm{corr}}&=\tfrac{1}{N\binom{C}{2}}\textstyle\sum_{i}\sum_{c<c'}\mathbb{I}[z_{i,c}\neq z_{i,c'}],\\
D_{\mathrm{out}}&=\tfrac{1}{N\binom{C}{2}}\textstyle\sum_{i}\sum_{c<c'}\mathbb{I}[\hat y_{i,c}\neq \hat y_{i,c'}],
\end{aligned}
\end{equation*}
where $\hat y_{i,c}$ is the predicted answer label for QA and the exact filtered candidate string for code. Uncertainty is quantified by percentile intervals from 1{,}000 bootstrap replicates that resample unique items with replacement while holding the configuration grid fixed (\Cref{alg:matched-metrics}; further diagnostics in \Cref{app:diagnostics}).}

\paragraph{Factors vs.\ Settings.}
Evaluation configurations for diffusion language models are naturally organized into two levels.
A \emph{factor} is a dimension of the evaluation protocol being varied
(e.g., numerical precision, Monte Carlo sampling, guidance scale, diffusion steps),
and a \emph{setting} is a concrete choice within a factor
(e.g., FP16 vs.\ BF16, or MC$=32$ vs.\ MC$=128$).
We \rev{contrast}
(i) differences \emph{between} factors \rev{with}
(ii) sensitivity \emph{within} a factor across its settings.

Let $i \in \{1,\ldots,k\}$ index factors and
$j \in \{1,\ldots,n_i\}$ index settings within factor $i$.
Let $x_{ij} \in \mathbb{R}$ denote the \rev{observed aggregate task score (accuracy or pass@1)} under setting $j$ of factor $i$\rev{; only factors that are varied for the task, and thus have at least two settings, are included}.
Let $n=\sum_{i=1}^k n_i$ be the total number of settings across all factors.

Define the factor mean and the grand mean as
\begin{equation}
\mu_i=\frac{1}{n_i}\sum_{j=1}^{n_i}x_{ij},
\qquad
\mu=\frac{1}{n}\sum_{i=1}^k\sum_{j=1}^{n_i}x_{ij}.
\label{eq:fva-means-main}
\end{equation}
We then define the between-factor and within-factor sums of squares:
\begin{equation}
\begin{aligned}
SSB&=\textstyle\sum_{i=1}^k n_i(\mu_i-\mu)^2,\\
SSW&=\textstyle\sum_{i=1}^k\sum_{j=1}^{n_i}(x_{ij}-\mu_i)^2.
\end{aligned}
\label{eq:fva-ss-main}
\end{equation}
Using the corresponding mean-square estimates,
\begin{equation}
\rev{MS_B}=\frac{SSB}{k-1},
\qquad
\rev{MS_W}=\frac{SSW}{n-k},
\label{eq:fva-var-main}
\end{equation}
we define \emph{Factor Variance Attribution (FVA)} as
\begin{equation}
\mathrm{FVA}
=
\frac{\rev{MS_B}}
     {\rev{MS_B+MS_W}}.
\label{eq:fva-main}
\end{equation}

FVA close to $1$ indicates that variability is dominated by
\emph{between-factor} differences,
whereas FVA close to $0$ indicates dominance of
\emph{within-factor} sensitivity.
\rev{When $MS_B+MS_W>0$, FVA lies in $[0,1]$; equal mean squares give $0.5$ rather than $0$. Under a common affine rescaling $x'=ax+b$ ($a\neq0$), both sums of squares scale by $a^2$ and FVA is unchanged, whereas factor-specific or nonlinear transformations can change it. With unequal $n_i$, the grand mean weights settings rather than factors. FVA therefore depends on the included factors, their ranges and numbers of settings, and the reference configuration; it is a finite-grid descriptive contrast rather than a population variance share. Because $x_{ij}$ is an aggregate score, FVA describes aggregate-score sensitivity on the evaluated grid and complements the matched-item measures above.}

\section{Single-Factor Analysis of Non-Determinism}

\subsection{Model Backbone}
\label{subsec:backbone}

We base our analysis on \textsc{LLaDA}~\cite{nie2025large} and its updated variant \textsc{\rev{LLaDA-1.5}}~\cite{zhu2025llada}, two \rev{checkpoints from the same LLaDA model family}~\cite{li2025survey} trained under a masked denoising formulation. Unlike autoregressive models~\cite{brown2020language,touvron2023llama,bi2024deepseek,bai2023qwen}, LLaDA generates text through an iterative diffusion process, predicting masked tokens in parallel at each step. Its multi-step and stochastic inference process makes model behavior inherently sensitive to inference-time configurations, providing a suitable setting for studying configuration-induced variability and sample-level non-determinism. Throughout this work, we use the publicly released 8B versions of LLaDA and \rev{LLaDA-1.5}. We focus on DLMs to encourage broader attention to non-determinism evaluation in diffusion-based generation, while additional results on autoregressive models (Qwen2.5-7B and \rev{LLaMA-2-7B}) are provided in Appendix~\ref{sec:ar} \rev{as cross-paradigm context}.

\subsection{Tasks and Datasets}
\label{subsec:tasks-datasets}

We evaluate DLMs on QA and code generation benchmarks to study variability across different task structures. For QA, we use \textsc{PIQA}~\cite{bisk2020piqa}, \textsc{WinoGrande}~\cite{sakaguchi2021winogrande}, and \textsc{ARC-Challenge}~\cite{clark2018think}; for code generation, we use \textsc{HumanEval}~\cite{chen2021evaluating} and \textsc{MBPP}~\cite{austin2021program}, where correctness is assessed by execution-based tests. Additional dataset details are in \Cref{sec:appendix-datasets}. Code is available at: \rev{\url{https://github.com/fangzy96/FVA}}.

\subsection{Factors and Configurations}
\label{subsec:factors}

We study evaluation non-determinism by varying a set of inference-time configuration factors,
covering both \emph{model-related} and \emph{system-related} sources of variability.
Our goal is not performance tuning, but to characterize how sensitive evaluation outcomes are
to different configuration choices.

\textbf{Model-related factors.}
These factors directly affect diffusion inference, including classifier-free guidance (CFG), diffusion steps, and Monte Carlo (MC) sampling. They influence stochastic generation dynamics and can induce variability even when dataset-level accuracy is similar. Since factor relevance differs across tasks, we evaluate them using task-appropriate configuration ranges rather than a uniform grid across benchmarks.

\textbf{System-related factors.}
We also study execution-level factors, including numerical precision, GPU type, and MC batch size. Although often treated as implementation details, these choices can introduce measurable sample-level non-determinism~\rev{\cite{yuan2025understanding}}. For each factor, we vary its settings while holding all others fixed \rev{by design} to a reference configuration to isolate its contribution to evaluation non-determinism rather than optimize peak accuracy. \rev{Details and documented exceptions} are provided in \Cref{sec:appendix-config}.

\subsection{Model-Related Factors}

We analyze sample-level non-determinism across two classes of factors: \emph{model-related} and \emph{system-related}. \Cref{fig:factor-std-grid} \rev{in \Cref{app:visualizations} shows dataset-level and cross-item Std for each factor across datasets and backbones.} \rev{Matched-item disagreement (\Cref{tab:qa-results-backbone,tab:generation-results-backbone}) shows that, for both factor classes, configuration changes alter the outcomes of a non-negligible fraction of items} even when dataset-level accuracy remains nearly unchanged. Model-related factors, including guidance strength, sampling dynamics, and convergence behavior, show pronounced sample-level non-determinism across tasks.

\textbf{Obs.~1.}
\rev{On QA, aggregate accuracy varies little across CFG and Monte Carlo settings ($s_a$ between $0.0045$ and $0.0455$; \Cref{tab:qa-results-backbone}),}
suggesting apparent robustness under conventional evaluation.
\rev{In contrast, matched-item measures reveal substantial changes: $D_{\mathrm{corr}}$ ranges from $0.0657$ to $0.2597$ for these factors; for example, LLaDA on PIQA under CFG has $s_a=0.0068$ but $D_{\mathrm{corr}}=0.0657$. On code generation, CFG and diffusion steps change both aggregate pass@1 and matched-item correctness ($s_a$ up to $0.1534$, $D_{\mathrm{corr}}$ between $0.1695$ and $0.2368$; \Cref{tab:generation-results-backbone}).}
This indicates that different model-related settings often lead to divergent predictions on the same inputs,
even when their aggregate accuracy is nearly identical.

\textbf{Obs.~2.}
Monte Carlo sampling is commonly used to reduce randomness by averaging over multiple stochastic trajectories.
However, \rev{changing the number of samples still alters many item-level outcomes.}
\rev{Across the evaluated MC budgets, matched-item disagreement remains the highest among QA factors ($D_{\mathrm{corr}}$ between $0.1327$ and $0.2597$; \Cref{tab:qa-results-backbone}).}
This suggests that variability is driven not only by stochastic noise,
but also by systematic differences in how individual inputs are resolved under different sampling configurations.

\subsection{System-Related Factors}

System-related factors arise from implementation and execution choices,
such as numerical precision, GPU type, and batch size.
Although often treated as secondary details,
these factors also have measurable effects on sample-level behavior.

\textbf{Obs.~1.}
Across numerical precision, GPU type, and batch size,
dataset-level accuracy again appears highly stable,
with negligible standard deviation across configurations.
\rev{Yet matched-item disagreement is nonzero: MC batch size yields $D_{\mathrm{corr}}$ between $0.0763$ and $0.1613$, comparable to CFG, whereas numerical precision and GPU type yield smaller but nonzero disagreement ($D_{\mathrm{corr}}$ between $0.0049$ and $0.0253$ on QA and between $0.0305$ and $0.0976$ on code).}
This demonstrates that sample-level non-determinism is not limited to changes in model logic,
but also arises \rev{under system-induced changes} during inference.

\textbf{Obs.~2.}
\rev{All 95\% bootstrap intervals for $D_{\mathrm{corr}}$ exclude zero, so these matched-item differences are unlikely to be an artifact of item sampling. The intervals quantify uncertainty over evaluation items only; whether repeated runs under an identical protocol reproduce the same outputs is a separate question that requires fixed-protocol repetitions.}
Together, these results suggest that configuration-induced sample-level non-determinism
is a structural property of \rev{inference in the evaluated LLaDA-family checkpoints},
rather than a consequence of isolated factors or random fluctuations.

\subsection{\rev{Fixed-Protocol Repetitions}}
\label{subsec:fixed-protocol}

\rev{The sweeps above deliberately change configurations. To assess strict non-determinism, we ran LLaDA-1.5 ten times per mode on PIQA and HumanEval under a frozen protocol (\Cref{app:new-experiments}): \emph{identical} launches reuse the evaluation seed, and \emph{seed-only} launches change only the seed. Identical launches were exactly reproducible on both tasks. Changing only the seed left HumanEval outputs unchanged (pass@1 $0.4146$) but altered PIQA predictions (label disagreement $0.1212$ [0.1129, 0.1305]; score SD $0.0082$) without a clear shift in mean accuracy ($0.0016$ [$-$0.0105, 0.0134]). Thus, with the seed fixed, no strict non-determinism was observed across the ten launches, whereas sensitivity to the seed value is task-dependent (\Cref{tab:fixed-protocol}).}

\section{Cross-Factor Variance Decomposition}
\label{sec:fva}

Building on the factor--setting distinction introduced earlier, we use FVA to \rev{contrast} factor identity \rev{with} setting-level sensitivity \rev{on the evaluated grid}. Unless otherwise specified, we use the same factors and settings as in the sample-level analysis. This section reports FVA results across datasets and backbones \rev{to describe how aggregate-score variation is distributed across the evaluated factors (reference-conditional first-order sensitivity, not causal attribution)}.

\subsection{Factor-Level Variance Analysis}
\label{subsec:fva-definition}

\begin{table}[]
\centering
\small
\revcolor
\resizebox{\columnwidth}{!}{%
\begin{tabular}{lcccc}
\toprule
Dataset & $MS_B$ & $MS_W$ & FVA [95\% CI] & $\eta^2$ \\
\midrule
\multicolumn{5}{c}{\textbf{LLaDA}} \\
\midrule
PIQA & 0.000728 & 0.000565 & 0.563\,[0.421,0.672] & 0.244 \\
WinoGrande & 0.000835 & 0.000716 & 0.538\,[0.337,0.684] & 0.226 \\
ARC-Challenge & 0.000325 & 0.000200 & 0.619\,[0.421,0.753] & 0.289 \\
HumanEval & 0.024057 & 0.006485 & 0.788\,[0.734,0.823] & 0.503 \\
MBPP & 0.036562 & 0.010764 & 0.773\,[0.744,0.795] & 0.481 \\
\midrule
\multicolumn{5}{c}{\textbf{LLaDA-1.5}} \\
\midrule
PIQA & 0.000902 & 0.000647 & 0.583\,[0.437,0.677] & 0.259 \\
WinoGrande & 0.000880 & 0.000505 & 0.635\,[0.465,0.740] & 0.304 \\
ARC-Challenge & 0.000616 & 0.000189 & 0.765\,[0.596,0.846] & 0.449 \\
HumanEval & 0.022138 & 0.011179 & 0.664\,[0.589,0.722] & 0.373 \\
MBPP & 0.012849 & 0.006823 & 0.653\,[0.589,0.698] & 0.361 \\
\bottomrule
\end{tabular}
}
\caption{\rev{Factor Variance Attribution (FVA) across datasets for two LLaDA-family checkpoints. For each dataset, only factors varied for that task enter the grid. We report the between- and within-factor mean squares ($MS_B$, $MS_W$), FVA with a 95\% unique-item bootstrap interval (1{,}000 replicates sharing item draws across all cells), and the observed-grid $\eta^2$. All values are conditional on the evaluated factors, settings, setting weights, and reference configuration.}}
\label{tab:fva-by-dataset-backbone}
\vspace{-15pt}
\end{table}

Aggregating across all datasets, we obtain
\rev{$MS_B=0.3157$, $MS_W=0.0302$, and $\mathrm{FVA}=0.913$}
for LLaDA, and
\rev{$MS_B=0.2153$, $MS_W=0.0241$, and $\mathrm{FVA}=0.899$}
for LLaDA-1.5.
\rev{Because the applicable factors differ between QA and code generation (e.g., diffusion steps are varied only for code), this pooled contrast partly reflects task composition and should not be read as the share of variability attributable to factor identity.}

\textbf{\rev{Obs.~1.}} \rev{Within every dataset, $MS_B$ exceeds $MS_W$ for both backbones (FVA between $0.538$ and $0.788$; \Cref{tab:fva-by-dataset-backbone}).} Nevertheless, within-factor sensitivity remains non-trivial, which is consistent with the \rev{large aggregate-score spread of diffusion steps and CFG on code generation ($s_a$ up to $0.1534$; \Cref{tab:generation-results-backbone})}.

\textbf{\rev{Obs.~2.}} \Cref{tab:fva-by-dataset-backbone} \rev{shows a task dependency that differs across backbones: for LLaDA, the contrast is higher on code generation ($0.788$, $0.773$) than on QA ($0.538$--$0.619$), whereas for LLaDA-1.5, ARC-Challenge has the highest value ($0.765$). Given the overlapping LLaDA-1.5 intervals and design sensitivity (\Cref{app:fva_sensitivity}), we do not interpret this relative contrast as an intrinsic task property.}

\textbf{\rev{Obs.~3.}} \rev{Backbone differences do not remove matched-item instability: the MC grids show similar $D_{\mathrm{corr}}$ for LLaDA ($0.1327$--$0.2574$) and LLaDA-1.5 ($0.1452$--$0.2597$; \Cref{tab:qa-results-backbone}), and \rev{diffusion steps remain} highly sensitive for both.}

These results suggest that stability-oriented evaluation should be both \emph{task-aware} and \emph{factor-aware}. \rev{On the factors varied for both task types (CFG, precision, and GPU type), code generation shows larger matched-item disagreement than QA and, except for LLaDA on MBPP under GPU type, larger aggregate-score spread (\Cref{tab:qa-results-backbone,tab:generation-results-backbone}).} \rev{Moving from LLaDA to LLaDA-1.5 does not eliminate evaluation non-determinism.} Appendix~\ref{sec:ar} \rev{reports aggregate-score variation for autoregressive models as cross-paradigm context.}

\subsection{\rev{Interaction and Remasking Checks}}
\label{subsec:interaction}

\rev{One-factor-at-a-time sweeps cannot capture interactions. On a $3\times3$ CFG ($0,1,2$) $\times$ Steps ($128,256,512$) grid for LLaDA-1.5 on HumanEval (\Cref{tab:cfg-steps-grid}), the marginal pass@1 ranges are $0.1951$ [0.1484, 0.2419] for CFG and $0.1545$ [0.1118, 0.1972] for Steps, whereas the interaction RMS is $0.0275$ [0.0171, 0.0422] and the largest absolute residual is $0.0447$ [0.0291, 0.0752]. The CFG-minus-Steps range difference, $0.0407$ [$-$0.0061, 0.0874], crosses zero, so we make no global ranking claim; other interactions, such as precision $\times$ hardware, remain untested. Replacing low-confidence remasking with random remasking lowered pass@1 by $0.1159$ at 128 steps and by $0.2073$ at 512 steps (difference-in-differences $-0.0915$ [$-$0.1707, $-$0.0061]; \Cref{tab:remasking}), so on this grid the policy effect depends on the number of steps. This broadens, but does not exhaust, the evaluated design.}

\paragraph{\rev{Practical recommendations.}}
\rev{We recommend separating fixed-protocol repetitions, system changes, and decoding sweeps; prespecifying the design; and judging material instability on absolute matched-item measures against a prespecified tolerance before localizing it with FVA (\Cref{app:checklist}).}

\section{Conclusion}
We showed that dataset-level metrics \rev{mask} non-determinism in \rev{the evaluated LLaDA-family} diffusion language models by aggregating sample-level behavior: configurations with similar aggregate performance can produce different predictions for the same inputs. To address this gap, we conducted single-factor analyses of model-related and system-related sources of variability, and introduced Factor Variance Attribution (FVA) \rev{as a finite-grid descriptive contrast of} between-factor and within-factor \rev{variation}. Our results show that \rev{configuration-induced non-determinism in the evaluated LLaDA-family checkpoints} is pervasive and structured, especially for code generation, highlighting the need for fine-grained, factor-aware evaluation beyond dataset-level metrics.

\section*{Limitations}
This work focuses on inference-time configuration variability and studies sample-level and factor-level effects on a set of standard benchmarks. Our analysis is scoped to evaluation-time behavior and does not consider training-time randomness or detailed semantic categorization of error types, which are complementary to the perspective developed here.

The proposed evaluation framework naturally extends to several directions. Applying sample-level and factor-level analysis to a broader range of tasks, including long-horizon reasoning and structured prediction, may reveal richer non-determinism patterns.
\rev{Our interaction analysis covers a single CFG $\times$ Steps grid on LLaDA-1.5/HumanEval, and the fixed-protocol repetitions cover one backbone, two tasks, and ten runs per mode; other interactions, such as precision $\times$ hardware, remain untested. In three historical grids, the harness batch size also varies within the sweep (\Cref{sec:appendix-config}).}
\rev{The empirical DLM evidence covers two checkpoints from the LLaDA family; independently developed DLM families remain to be tested, and the autoregressive results in \Cref{sec:ar} provide only cross-paradigm context.} Beyond diffusion language models, the same methodology can be applied to a wider set of model backbones and generative paradigms to assess the generality of configuration-induced variability.
Finally, jointly studying training-time and inference-time sources of variability may further improve the robustness and interpretability of evaluation protocols.

\section*{Acknowledgments}
\rev{This work made use of the High Performance Computing Resource in the Core Facility for Advanced Research Computing at Case Western Reserve University. This work was supported in part by NSF CCF-2200255, NSF CCF-2006780, NSF IIS-2027667, NIH U01AG073323, NIH R01HG009658, NIH 1R01HL159170-01A1, as well as by the Clinical and Translational Science Collaborative of Cleveland, which is funded by the National Institutes of Health, National Center for Advancing Translational Sciences, through the Clinical and Translational Science Award UL1TR002548.}

\bibliography{custom}

\appendix
\renewcommand{\topfraction}{0.9}\renewcommand{\dbltopfraction}{0.9}
\renewcommand{\floatpagefraction}{0.6}\renewcommand{\dblfloatpagefraction}{0.6}
\renewcommand{\textfraction}{0.1}\setcounter{topnumber}{3}\setcounter{dbltopnumber}{3}
\makeatletter
\setlength{\@fptop}{0pt}\setlength{\@fpbot}{0pt plus 1fil}
\setlength{\@dblfptop}{0pt}\setlength{\@dblfpbot}{0pt plus 1fil}
\makeatother

\section{Detailed Description of Datasets}
\label{sec:appendix-datasets}

This section provides detailed descriptions of the datasets used in our experiments.

\begin{itemize}
  \item \textbf{PIQA}~\cite{bisk2020piqa}%
  \footnote{\url{https://huggingface.co/datasets/ybisk/piqa}}:
  A multiple-choice question answering benchmark focused on physical commonsense reasoning. Each instance presents a goal description and two candidate solutions, only one of which is physically plausible.

  \item \textbf{WinoGrande}~\cite{sakaguchi2021winogrande}%
  \footnote{\url{https://huggingface.co/datasets/allenai/winogrande}}:
  A large-scale adversarial dataset for commonsense reasoning based on pronoun resolution, constructed to reduce annotation artifacts and spurious correlations.

  \item \textbf{ARC-Challenge}~\cite{clark2018think}%
  \footnote{\url{https://huggingface.co/datasets/allenai/ai2_arc}}:
  The more difficult subset of the AI2 Reasoning Challenge, consisting of grade-school science exam questions that typically require multi-step reasoning beyond surface pattern matching.

  \item \textbf{HumanEval}~\cite{chen2021evaluating}%
  \footnote{\url{https://huggingface.co/datasets/openai/openai_humaneval}}:
  A code generation benchmark composed of Python programming problems. Model outputs are evaluated based on functional correctness using hidden unit tests.

  \item \textbf{MBPP}~\cite{austin2021program}%
  \footnote{\url{https://huggingface.co/datasets/google-research-datasets/mbpp}}:
  The Mostly Basic Programming Problems dataset, which includes a larger collection of programming tasks with varying difficulty levels. Similar to HumanEval, evaluation is based on execution against unit tests.
\end{itemize}

\section{Configuration Details}
\label{sec:appendix-config}

This section provides additional details on the inference-time configurations considered in our experiments, and places them in the context of prior findings on hyperparameter and system-level effects.
Previous work has shown that hyperparameters and execution-related choices can substantially influence model behavior, often in ways that are difficult to predict or diagnose~\cite{sculley2015hidden,arora2024optimizing,donato2025studying}.
In large-scale models, sources of variability such as random seeds, numerical precision, hardware platforms, and software stacks have been shown to introduce measurable differences in inference outcomes, even when training and evaluation pipelines are otherwise identical~\cite{shanmugavelu2024impacts,yuan2025understanding}.

Motivated by these observations, our experimental design explicitly varies a set of inference-time configuration factors spanning both model-related and system-related sources of variability.
Model-related factors include classifier-free guidance (CFG) scale, diffusion steps, and Monte Carlo (MC) sampling, which directly affect the stochastic diffusion inference process.
System-related factors include numerical precision, GPU type, and the batch size used for MC sampling, which influence numerical execution and parallelism during inference.

For each factor, we evaluate multiple concrete settings while holding all other factors fixed \rev{by design}.
This controlled design \rev{is intended} to isolate the effect of individual factors and to compare variability arising from differences between factors versus sensitivity to specific settings within the same factor.
Unlike most prior studies that examine individual configuration choices in isolation or focus on aggregate performance differences across runs, our setup enables a unified, variance-based \rev{comparison of multiple configuration classes on the evaluated grid}.
Detailed lists of factors, settings, and implementation choices are provided below.

\paragraph{General Protocol.}
All experiments focus on inference-time variability.
For each factor under study, we vary its settings while holding all other factors
fixed to a common reference configuration \rev{by design}.
This one-factor-at-a-time design \rev{is intended to isolate} the effect of each factor within a task,
rather than enabling direct cross-task comparison across heterogeneous factor sets.
No hyperparameter tuning is performed for performance optimization;
all configurations are evaluated under the same protocol
to assess stability rather than peak accuracy.

\subsection{Model-Related Factors}
\emph{Classifier-Free Guidance (CFG).}
We vary the guidance scale applied during diffusion inference, which controls the trade-off between conditional fidelity and generation diversity. All other inference parameters are held fixed \rev{by design} when varying CFG. We evaluate guidance scales of 0, 0.5, 1.0, 1.5, and 2.0.

\emph{Diffusion Steps.}
We vary the total number of denoising steps used during inference. This affects convergence behavior and generation quality but does not alter the model parameters. For \rev{code} generation datasets, we evaluate step counts of 64, 128, 256, and 512. In addition, for LLaDA we also include 1024 steps, while this setting is not evaluated for LLaDA-1.5, as its original paper does not recommend using more than 512 steps.

\emph{Monte Carlo (MC) Sampling.}
MC sampling refers to repeating the diffusion inference process multiple times with different random seeds and aggregating predictions. The MC setting controls the number of independent samples used for aggregation, while the aggregation rule itself is fixed across experiments. For question answering datasets, we evaluate MC sizes of 8, 16, 32, 64, 128, and 256.

\subsection{System-Related Factors}
\emph{Numerical Precision.}
We evaluate inference under three numerical precision settings (FP16, BF16, and INT8), while keeping model weights and inference logic unchanged. \rev{The saved configurations of these sweeps do not jointly record the weight and confidence-computation dtypes for every cell, so we do not retrospectively claim that all softmax computations used FP32; the experiments in \Cref{app:new-experiments} record both dtypes explicitly.}

\emph{GPU Type.}
We compare inference results across two GPU architectures, using four H100 GPUs and four A100 GPUs. \rev{The intended configuration specifies CUDA~12.3 and Python~3.10.18, but the saved result files do not record these fields for every run. These sweeps probe system-induced changes, not same-condition kernel non-determinism.}

\emph{Batch Size for MC Sampling.}
Batch size refers to the number of Monte Carlo (MC) samples processed concurrently during inference. This parameter affects execution order and numerical behavior but does not change the total number of MC samples aggregated for prediction. For question answering datasets, we evaluate batch sizes of \rev{1, 4, 8, 16, and 32}. 

\paragraph{Reproducibility.}
Unless explicitly varied as part of a factor, random seeds, software versions, and decoding logic are held fixed \rev{by design. However, the historical result files do not jointly record every execution field (e.g., seed, confidence dtype, and remasking policy) for every cell. We report only documented provenance, do not infer missing fields, and treat the sweeps as descriptive matched-output evidence of configuration sensitivity. The experiments in \Cref{app:new-experiments} follow a fully recorded protocol. In three historical grids (LLaDA on HumanEval under CFG and under diffusion steps, and LLaDA on MBPP under GPU type), the recorded harness batch size also differs across cells (1 versus 32), so differences in these grids cannot be attributed to the named factor alone.}

\paragraph{\rev{Matched-item metric computation.}}
\rev{\Cref{alg:matched-metrics} summarizes how the matched-item quantities in \Cref{tab:qa-results-backbone,tab:generation-results-backbone} and their bootstrap intervals are computed for one model--dataset--factor grid.}

\begin{algorithm}[t]
\revcolor
\caption{\rev{Matched-item metrics for one factor grid}}
\label{alg:matched-metrics}
\begin{algorithmic}[1]
\Require configurations $\mathcal C$ of one factor; outputs of each $c\in\mathcal C$ keyed by item
\State $\mathcal I\gets\bigcap_{c\in\mathcal C}\mathcal I_c$ \Comment{match by document hash, else dataset and document ID}
\State build $Z=(z_{i,c})$ and $\hat Y=(\hat y_{i,c})$ for $i\in\mathcal I$, $c\in\mathcal C$; $N\gets|\mathcal I|$, $C\gets|\mathcal C|$
\State $a_c\gets\frac1N\sum_i z_{i,c}$; $\bar a, s_a\gets$ mean and SD (ddof$=1$) of $\{a_c\}$
\State $V_{\mathrm{item}}\gets\frac1N\sum_i\mathrm{Var}_c(z_{i,c})$ \Comment{ddof$=0$}
\State $\mathrm{Std}_x\gets$ SD (ddof$=1$) of $\{\bar z_i\}$, where $\bar z_i=\frac1C\sum_c z_{i,c}$
\State $D_{\mathrm{corr}}, D_{\mathrm{out}}\gets$ mean fraction of unequal pairs $c<c'$ in $Z$, $\hat Y$
\For{$b=1,\dots,1000$}
  \State draw $N$ items from $\mathcal I$ with replacement, keeping each item's full rows of $Z$ and $\hat Y$
  \State recompute $V_{\mathrm{item}}^{(b)}, D_{\mathrm{corr}}^{(b)}, D_{\mathrm{out}}^{(b)}$
\EndFor
\State \Return point estimates and the 2.5th/97.5th percentiles of the replicates
\end{algorithmic}
\end{algorithm}

\section{Additional Visualizations}
\label{app:visualizations}

\begin{figure*}[t]
    \centering
    \includegraphics[width=1.0\textwidth]{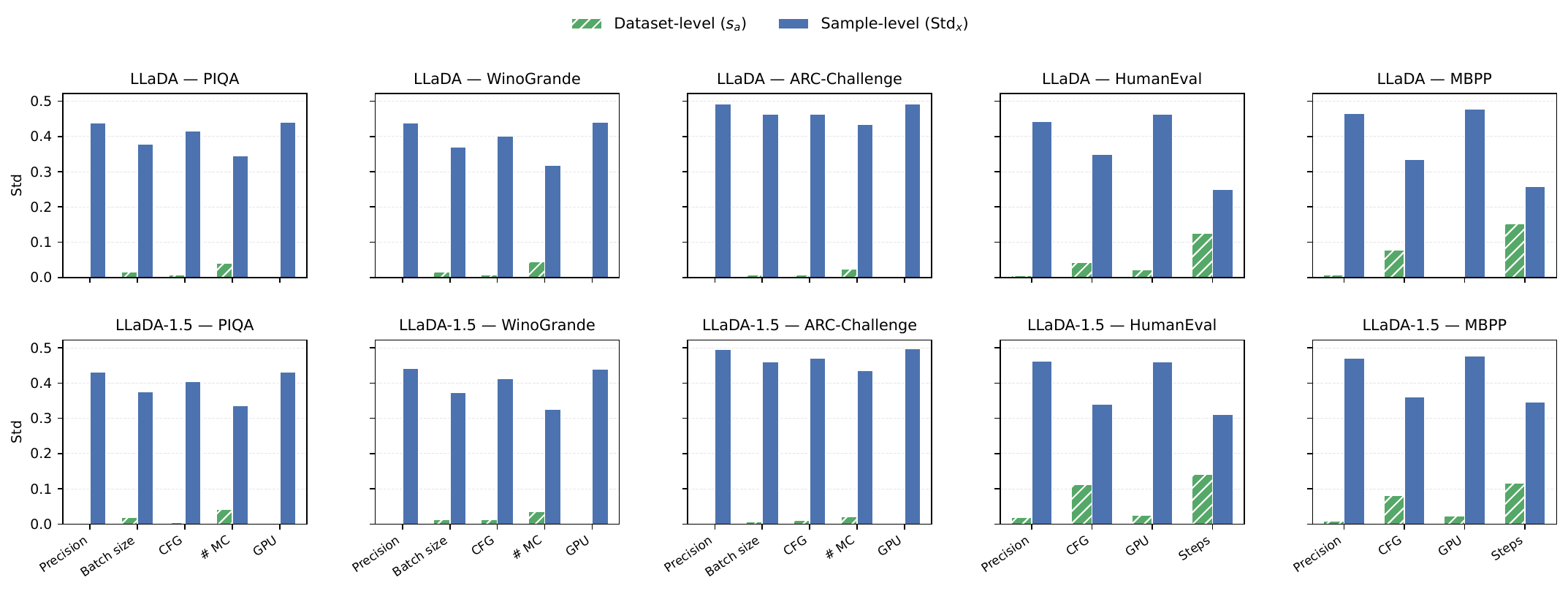}
    \caption{
    \rev{Dataset-level and sample-level Std for each factor, dataset, and backbone. Dataset-level bars are the standard deviation of aggregate scores across the evaluated configurations ($s_a$); sample-level bars are the cross-item standard deviation of configuration-averaged correctness ($\mathrm{Std}_x$; \Cref{sec:evaluation-paradigm}), which reflects heterogeneity across items rather than within-item change across configurations.}
    Results are shown for both LLaDA and LLaDA-1.5 on question answering and code generation tasks.
    }
    \label{fig:factor-std-grid}
\end{figure*}

\begin{figure*}[t]
    \centering
    \includegraphics[width=1.0\textwidth]{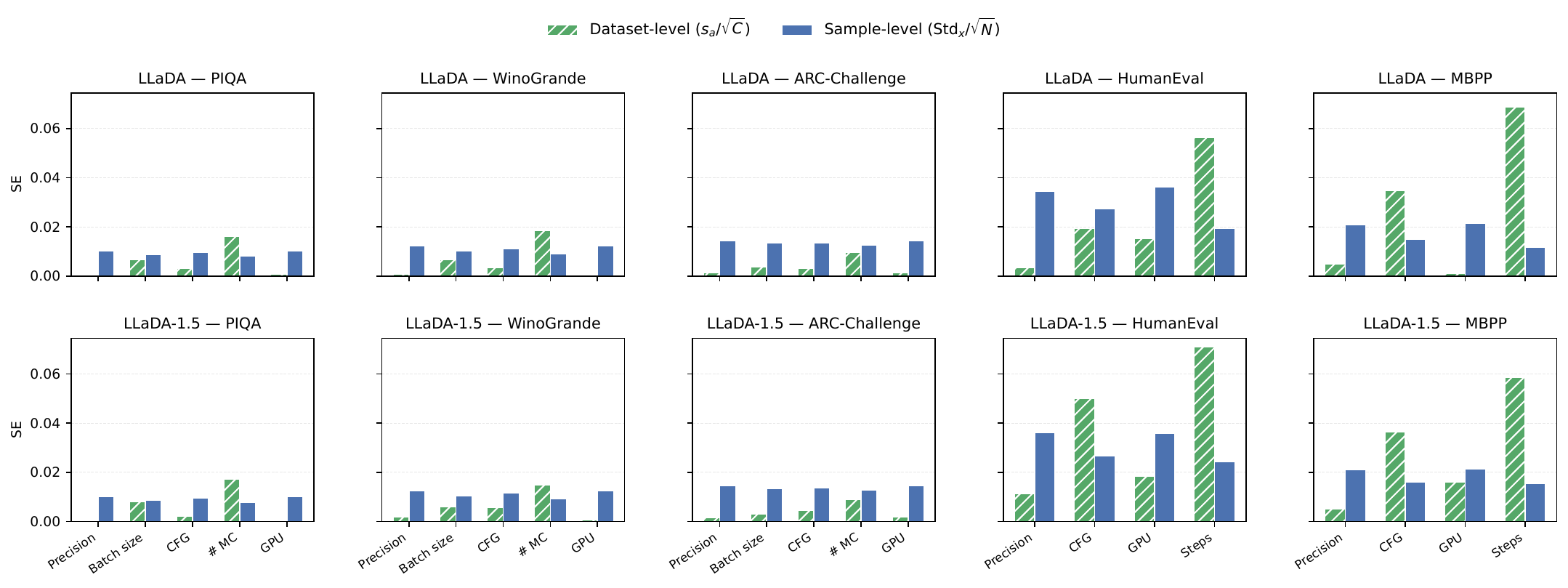}
    \caption{
    \rev{Dataset-level and sample-level SE for each factor, dataset, and backbone, computed as $s_a/\sqrt{C}$ and $\mathrm{Std}_x/\sqrt{N}$, respectively. Both are descriptive summaries of the quantities in \Cref{fig:factor-std-grid}; they do not quantify the reliability of a factor effect, matched-item instability, or fixed-protocol reproducibility.}
    }
    \label{fig:factor-se-grid}
\end{figure*}

\rev{\Cref{fig:factor-std-grid} shows the dataset-level and cross-item standard deviations for each factor.}

\rev{\Cref{fig:factor-se-grid} reports the corresponding standard errors. Because they divide the same standard deviations by the square root of the number of configurations or items, they are descriptive and are not used as evidence of matched-item instability; the bootstrap intervals in \Cref{tab:qa-results-backbone,tab:generation-results-backbone} quantify item-sampling uncertainty instead.}

\section{Case Studies of Sample-Level Non-Determinism}
\label{sec:case-studies}

Table~\ref{tab:sample-level-disagreement-mc} presents additional qualitative examples illustrating
sample-level non-determinism induced by Monte Carlo (MC) sampling.
For both questions, varying the number of MC samples leads to different predicted answers for the same input,
despite all configurations being evaluated under the same model backbone and task setting.
Importantly, these configurations \rev{contribute identically to dataset-level accuracy on these items}, yet exhibit markedly different behaviors at the sample level.

In the first example, predictions fluctuate across multiple incorrect options as the MC size increases,
indicating that the model’s uncertainty is resolved differently under different sampling budgets.
In the second example, predictions alternate between two answer choices, neither of which matches the ground truth,
demonstrating that correctness remains consistently incorrect while the specific error mode changes.
Such behavior highlights that sample-level non-determinism can arise even in the absence of correctness flips,
and that different inference-time configurations may correspond to qualitatively different failure paths.

These case studies reinforce the limitation of dataset-level evaluation:
aggregate metrics cannot reveal which inputs are non-deterministic, nor how model errors vary across configurations.
By contrast, sample-level analysis exposes both correctness variability and error-mode diversity,
providing critical insight into the reliability of diffusion language model predictions under stochastic inference.

\begin{table*}[t]
\centering
\small
\resizebox{1.0\textwidth}{!}{%
\begin{tabular}{p{6.8cm} p{6.8cm} c c c c c c p{1.6cm}}
\toprule
Question & Answer Options
& MC=8 & MC=16 & MC=32 & MC=64 & MC=128 & MC=256 & Ground Truth \\
\midrule

Which statement best explains why the lower mantle of Earth is much more rigid and dense than the upper mantle?
&
A: The lower mantle is older \newline
B: The lower mantle is cooler \newline
C: The lower mantle is under more pressure \newline
D: The lower mantle is farther from the core
&
D & A & B & B & B & B & C \\

Which statement is an opinion about the heart?
&
A: The heart moves blood through the body \newline
B: The heart helps perform athletic activities \newline
C: The heart rate changes during the day \newline
D: The heart is composed of different types of cells
&
A & A & D & D & D & A & B \\

\bottomrule
\end{tabular}
}
\caption{Additional examples of sample-level non-determinism under different Monte Carlo (MC) sampling \rev{budgets}.}
\label{tab:sample-level-disagreement-mc}
\end{table*}

\begin{table*}[t]
\centering
\resizebox{\textwidth}{!}{%
\begin{tblr}{
  colspec = {l l c c c c c c c c c},
  cell{1}{1} = {r=2}{m},
  cell{1}{2} = {r=2}{m},
  cell{1}{3} = {c=3}{c},
  cell{1}{6} = {c=3}{c},
  cell{1}{9} = {c=3}{c},
  cell{3}{1}  = {r=8}{m},
  cell{11}{1} = {r=8}{m},
  hline{1,3,11,19} = {-}{},
}
\textbf{Backbone} & \textbf{Factor}
& \textbf{PIQA} & & 
& \textbf{WinoGrande} & & 
& \textbf{ARC-Challenge} & & \\
& 
& Accuracy & Dataset-level & Sample-level
& Accuracy & Dataset-level & Sample-level
& Accuracy & Dataset-level & Sample-level \\

\rotatebox{90}{\textbf{LLaMA-2-7B}}
& Precision (Std)
& $0.7818$ & $0.0005$ & $0.4081$
& $0.6967$ & $0.0051$ & $0.4442$
& $0.4266$ & $0.0034$ & $0.4863$ \\

& Precision (SE)
& $0.7818$ & $0.0003$ & $0.0095$
& $0.6967$ & $0.0029$ & $0.0125$
& $0.4266$ & $0.0020$ & $0.0142$ \\

& Batch size (Std)
& $0.7817$ & $0.0007$ & $0.4127$
& $0.6938$ & $0.0021$ & $0.4565$
& $0.4232$ & $0.0000$ & $0.4936$ \\

& Batch size (SE)
& $0.7817$ & $0.0003$ & $0.0096$
& $0.6938$ & $0.0011$ & $0.0128$
& $0.4232$ & $0.0000$ & $0.0144$ \\

& Temperature (Std)
& $0.7824$ & $0.0000$ & $0.4127$
& $0.6930$ & $0.0000$ & $0.4614$
& $0.4232$ & $0.0000$ & $0.4943$ \\

& Temperature (SE)
& $0.7824$ & $0.0000$ & $0.0096$
& $0.6930$ & $0.0000$ & $0.0130$
& $0.4232$ & $0.0000$ & $0.0144$ \\

& GPUs (Std)
& $0.7802$ & $0.0031$ & $0.4123$
& $0.6946$ & $0.0022$ & $0.4556$
& $0.4249$ & $0.0024$ & $0.4924$ \\

& GPUs (SE)
& $0.7802$ & $0.0022$ & $0.0096$
& $0.6946$ & $0.0016$ & $0.0128$
& $0.4249$ & $0.0017$ & $0.0144$ \\

\rotatebox{90}{\textbf{Qwen2.5-7B}}
& Precision (Std)
& $0.7876$ & $0.0006$ & $0.4042$
& $0.7285$ & $0.0041$ & $0.4301$
& $0.4807$ & $0.0025$ & $0.4908$ \\

& Precision (SE)
& $0.7876$ & $0.0004$ & $0.0094$
& $0.7285$ & $0.0024$ & $0.0121$
& $0.4807$ & $0.0014$ & $0.0143$ \\

& Batch size (Std)
& $0.7871$ & $0.0014$ & $0.4072$
& $0.7301$ & $0.0011$ & $0.4395$
& $0.4814$ & $0.0013$ & $0.4963$ \\

& Batch size (SE)
& $0.7871$ & $0.0007$ & $0.0095$
& $0.7301$ & $0.0006$ & $0.0123$
& $0.4814$ & $0.0006$ & $0.0145$ \\

& Temperature (Std)
& $0.7873$ & $0.0000$ & $0.4094$
& $0.7309$ & $0.0000$ & $0.4437$
& $0.4821$ & $0.0000$ & $0.4999$ \\

& Temperature (SE)
& $0.7873$ & $0.0000$ & $0.0095$
& $0.7309$ & $0.0000$ & $0.0125$
& $0.4821$ & $0.0000$ & $0.0146$ \\

& GPUs (Std)
& $0.7873$ & $0.0000$ & $0.4077$
& $0.7313$ & $0.0006$ & $0.4388$
& $0.4804$ & $0.0024$ & $0.4973$ \\

& GPUs (SE)
& $0.7873$ & $0.0000$ & $0.0095$
& $0.7313$ & $0.0004$ & $0.0123$
& $0.4804$ & $0.0017$ & $0.0145$ \\

\end{tblr}
}
\caption{
Dataset-level and sample-level evaluation results on question answering datasets
for autoregressive language model backbones.
\rev{For each factor, we report mean accuracy, the standard deviation of aggregate accuracy across configurations (dataset-level Std), and the cross-item standard deviation of configuration-averaged correctness (sample-level Std, $\mathrm{Std}_x$ in \Cref{sec:evaluation-paradigm}), which reflects heterogeneity across items rather than within-item change. SE rows divide these by the square root of the number of configurations or items and are descriptive.}
}
\label{tab:qa-results-ar}
\end{table*}

\begin{table*}[t]
\small
\centering
\resizebox{0.8\textwidth}{!}{%
\begin{tblr}{
  colspec = {l l c c c c c c},
  cell{1}{1} = {r=2}{m},
  cell{1}{2} = {r=2}{m},
  cell{1}{3} = {c=3}{c},
  cell{1}{6} = {c=3}{c},
  cell{3}{1} = {r=8}{m},
  cell{11}{1} = {r=8}{m},
  hline{1,3,11,19} = {-}{},
}
\textbf{Backbone} & \textbf{Factor} 
& \textbf{HumanEval} & & 
& \textbf{MBPP} & & \\
& 
& Pass@1 & Dataset-level & Sample-level 
& Pass@1 & Dataset-level & Sample-level \\

\rotatebox{90}{\textbf{LLaMA-2-7B}}
& Precision (Std) & $0.0650$ & $0.0254$ & $0.1517$ & $0.1940$ & $0.0140$ & $0.3227$ \\
& Precision (SE)  & $0.0650$ & $0.0147$ & $0.0118$ & $0.1940$ & $0.0081$ & $0.0144$ \\
& Batch size (Std) & $0.0549$ & $0.0122$ & $0.1358$ & $0.1935$ & $0.0090$ & $0.3172$ \\
& Batch size (SE)  & $0.0549$ & $0.0061$ & $0.0106$ & $0.1935$ & $0.0045$ & $0.0142$ \\
& Temperature (Std) & $0.0732$ & $0.0198$ & $0.1292$ & $0.2076$ & $0.0250$ & $0.3224$ \\
& Temperature (SE)  & $0.0732$ & $0.0088$ & $0.0101$ & $0.2076$ & $0.0112$ & $0.0144$ \\
& GPUs (Std) & $0.0549$ & $0.0259$ & $0.1663$ & $0.2020$ & $0.0028$ & $0.3513$ \\
& GPUs (SE)  & $0.0549$ & $0.0183$ & $0.0130$ & $0.2020$ & $0.0020$ & $0.0157$ \\

\rotatebox{90}{\textbf{Qwen2.5-7B}}
& Precision (Std) 
& $0.4695$ & $0.0106$ & $0.4075$ 
& $0.6000$ & $0.0312$ & $0.4268$ \\

& Precision (SE)  
& $0.4695$ & $0.0061$ & $0.0318$ 
& $0.6000$ & $0.0180$ & $0.0191$ \\

& Batch size (Std) 
& $0.4665$ & $0.0061$ & $0.3882$ 
& $0.6125$ & $0.0180$ & $0.4139$ \\

& Batch size (SE)  
& $0.4665$ & $0.0030$ & $0.0303$ 
& $0.6125$ & $0.0090$ & $0.0185$ \\

& Temperature (Std) 
& $0.4439$ & $0.0893$ & $0.3766$ 
& $0.5736$ & $0.0839$ & $0.3959$ \\

& Temperature (SE)  
& $0.4439$ & $0.0400$ & $0.0294$ 
& $0.5736$ & $0.0375$ & $0.0177$ \\

& GPUs (Std) 
& $0.4817$ & $0.0259$ & $0.4322$ 
& $0.6100$ & $0.0113$ & $0.4362$ \\

& GPUs (SE)  
& $0.4817$ & $0.0183$ & $0.0337$ 
& $0.6100$ & $0.0080$ & $0.0195$ \\

\end{tblr}
}
\caption{
Dataset-level and sample-level evaluation results on \rev{code} generation datasets
for autoregressive language model backbones.
Within each backbone, results follow the same evaluation protocol as in
Table~\ref{tab:qa-results-ar}.
\rev{For HumanEval, the LLaMA-2-7B rows use CodeLLaMA-2-7B.}
}
\label{tab:generation-results-ar}
\end{table*}

\begin{table}[t]
\centering
\small
\resizebox{0.9\columnwidth}{!}{%
\begin{tabular}{lccc}
\toprule
Dataset
& $\sigma^2_{\mathrm{between}}$
& $\sigma^2_{\mathrm{within}}$
& FVA \\
\midrule
\multicolumn{4}{c}{\textbf{LLaMA-2-7B}} \\
\midrule
PIQA & 0.000002 & 0.000001 & 0.664 \\
WinoGrande & 0.000009 & 0.000007 & 0.559 \\
ARC-Challenge & 0.000009 & 0.000003 & 0.759 \\
HumanEval & 0.000307 & 0.000397 & 0.436 \\
MBPP & 0.000191 & 0.000314 & 0.378 \\
\midrule
\multicolumn{4}{c}{\textbf{Qwen2.5-7B}} \\
\midrule
PIQA & 0.000000 & 0.000001 & 0.177 \\
WinoGrande & 0.000004 & 0.000004 & 0.539 \\
ARC-Challenge & 0.000002 & 0.000002 & 0.468 \\
HumanEval & 0.000885 & 0.003293 & 0.212 \\
MBPP & 0.001345 & 0.003120 & 0.301 \\
\bottomrule
\end{tabular}
}
\caption{Factor Variance Attribution (FVA) across datasets for autoregressive language model backbones.
For each backbone, we report \rev{the between- and within-factor mean squares ($MS_B$, $MS_W$; columns labeled $\sigma^2_{\mathrm{between}}$ and $\sigma^2_{\mathrm{within}}$)} and the resulting FVA. \rev{Values are relative contrasts conditional on the evaluated grid.}}
\label{tab:fva-by-dataset-ar}
\end{table}

\begin{table}[t]
\centering
\small
\resizebox{1.0\columnwidth}{!}{%
\begin{tabular}{lccc|ccc}
\toprule
& \multicolumn{3}{c}{\textbf{LLaMA-2-7B}} & \multicolumn{3}{c}{\textbf{Qwen2.5-7B}} \\
Factor
& Mean & Std & SE
& Mean & Std & SE \\
\midrule
Batch
& 0.4294 & 0.2866 & 0.0641
& 0.6155 & 0.1322 & 0.0296 \\

Temperature
& 0.4359 & 0.2779 & 0.0556
& 0.6035 & 0.1466 & 0.0293 \\

GPU
& 0.4313 & 0.2928 & 0.0926
& 0.6181 & 0.1329 & 0.0420 \\

Precision
& 0.4328 & 0.2870 & 0.0741
& 0.6133 & 0.1332 & 0.0344 \\

\bottomrule
\end{tabular}
}
\caption{Factor-level summary statistics aggregated across datasets for autoregressive language model backbones.
For each factor, we report the mean score, standard deviation (Std), and standard error (SE) across settings.
\rev{Because cells are pooled across datasets, Std mixes between-dataset differences with within-factor sensitivity, and SE is a descriptive summary rather than a sampling interval.}}
\label{tab:factor-summary-ar}
\end{table}

\section{Non-Determinism Analysis on Autoregressive Language Models}
\label{sec:ar}

To complement our study on diffusion language models, we extend our analysis to autoregressive language models, including LLaMA-2-7B~\cite{touvron2023llama} and Qwen2.5-7B~\rev{\cite{qwen2025qwen25technicalreport}}\rev{, as cross-paradigm context}. Unlike diffusion models, autoregressive models generate tokens sequentially and are often considered more stable under fixed decoding configurations. (For \rev{HumanEval}, we use CodeLLaMA-2-7B\rev{~\cite{roziere2024codellamaopenfoundation}} instead of LLaMA-2-7B due to known whitespace handling issues in the tokenizer, which can affect evaluation.)

We follow the same evaluation protocol as in the main paper, measuring both dataset-level performance and sample-level variability across different inference factors. Tables~\ref{tab:qa-results-ar} and~\ref{tab:generation-results-ar} show that dataset-level variability is relatively small for most factors, especially for QA tasks under deterministic decoding settings. 
\rev{The sample-level columns of these tables use the cross-item reduction ($\mathrm{Std}_x$ in \Cref{sec:evaluation-paradigm}) and reflect heterogeneity across items; on their own, they do not show that individual predictions change across settings. Matched-item measures were not computed for these sweeps, so we use them only as cross-paradigm context, not as matched-item evidence or as validation on another DLM family.}

The Factor Variance Attribution results in Table~\ref{tab:fva-by-dataset-ar} provide a complementary view. Although the absolute variance is very small compared to diffusion models, \rev{QA tasks tend to have higher FVA values than code generation tasks. Because FVA is a relative, grid-dependent contrast, these values must be read together with the small absolute variation.}
Table~\ref{tab:factor-summary-ar} further shows that mean performance is very similar across factors, with only small differences in aggregate scores. \rev{This table pools cells across datasets and is descriptive.}

\rev{These sweeps show that aggregate scores of autoregressive models also vary across configurations: for example, the Qwen2.5 temperature sweep yields an aggregate-score SD of $0.0893$ on HumanEval and $0.0839$ on MBPP (\Cref{tab:generation-results-ar}). They do not substitute for validation on an independently developed DLM family and are not corrected matched-item evidence.}

\begin{table}[t]
\centering
\small
\revcolor
\resizebox{\columnwidth}{!}{%
\begin{tabular}{lccccc}
\toprule
\textbf{Dataset} & \textbf{Base} & \textbf{Alt} & $\boldsymbol{\Delta}$\textbf{FVA} & $\boldsymbol{\Delta\eta^2}$ & \textbf{Cells} \\
\midrule
PIQA & 0.5630 & 0.5669 & +0.0039 & $-$0.0081 & 21/22 \\
WinoGrande & 0.5382 & 0.5323 & $-$0.0059 & $-$0.0041 & 21/21 \\
ARC-Challenge & 0.6190 & 0.5986 & $-$0.0203 & $-$0.0290 & 21/22 \\
HumanEval & 0.7877 & 0.7972 & +0.0095 & +0.0145 & 15/15 \\
MBPP & 0.7726 & 0.7507 & $-$0.0218 & $-$0.0299 & 15/15 \\
\bottomrule
\end{tabular}
}
\caption{\rev{Limited sensitivity check of FVA to factor ranges and value selections (LLaDA). ``Base'' is the evaluated grid; ``Alt'' expands factor ranges (e.g., scaling CFG) and uses alternative values (e.g., diffusion steps). $\Delta$ is Alt $-$ Base for FVA and for observed-grid $\eta^2$; the last column gives the number of factor--setting cells. For ARC-Challenge and PIQA, the cell count also changes, so the difference combines range/value and setting-weight changes. No independent LLaDA-1.5 Alt grid is available.}}
\label{tab:fva_robustness}
\end{table}

\section{Sensitivity Analysis of FVA to Factor Ranges and Value Choices}
\label{app:fva_sensitivity}

To further examine the \rev{design sensitivity} of FVA, we perform a sensitivity analysis by varying both factor ranges and specific value selections.
Concretely, we compare the original configuration (\rev{``base settings''}) with an alternative configuration where:
(1) factor ranges are expanded (e.g., scaling CFG values by a factor of 2), and 
(2) different discrete values are chosen within the same range (e.g., alternative diffusion step counts). Table~\ref{tab:fva_robustness} summarizes the results. \rev{\Cref{tab:factor-summary-backbone} additionally gives the pooled factor-level summary of aggregate scores.}

\textbf{\rev{Obs.~1.}}
\rev{Across the five LLaDA datasets, the Base--Alt changes are small (largest absolute change $0.0218$ in FVA and $0.0299$ in $\eta^2$, both on MBPP) and show no consistent direction.}

\textbf{\rev{Obs.~2.}}
\rev{This is a limited perturbation rather than a general robustness test: it covers one backbone and two grids, and for ARC-Challenge and PIQA it also changes the number of cells. It does not establish insensitivity to omitted factors, other ranges, or interactions.}

\textbf{\rev{Obs.~3.}}
\rev{\Cref{tab:fva-design-sensitivity} shows that FVA depends on how settings and factors are weighted: equal-factor summaries range from $0.262$ to $0.569$, equal-level subsampling medians from $0.418$ to $0.682$, and leave-one-factor-out values from $0.196$ to $0.823$. Factor rankings are stable under equal-level subsampling in 9 of 10 cases (median $\rho\geq0.80$), with LLaDA-1.5/ARC-Challenge as the clear exception ($\rho=0.05$).}

\rev{These results support FVA as a descriptive summary of the evaluated grid, to be reported together with its design (factors, ranges, setting counts, weights, and reference configuration) and with absolute matched-item measures. They do not establish population-level variance attribution, and we did not fit a mixed-effects variance-component model because the factors and settings are deliberately chosen rather than sampled.}

\begin{table}[t]
\centering
\revcolor
\small
\resizebox{0.95\columnwidth}{!}{%
\begin{tabular}{lccc|ccc}
\toprule
& \multicolumn{3}{c}{\textbf{LLaDA}} & \multicolumn{3}{c}{\textbf{LLaDA-1.5}} \\
Factor
& Mean & Std & SE
& Mean & Std & SE \\
\midrule
Batch
& 0.6348 & 0.1453 & 0.0375
& 0.6640 & 0.0975 & 0.0252 \\
CFG
& 0.4886 & 0.2107 & 0.0421
& 0.5098 & 0.2095 & 0.0419 \\
GPU
& 0.5258 & 0.1828 & 0.0578
& 0.5579 & 0.1658 & 0.0524 \\
MC
& 0.6119 & 0.1410 & 0.0332
& 0.6382 & 0.0998 & 0.0235 \\
Precision
& 0.5257 & 0.1822 & 0.0470
& 0.5585 & 0.1572 & 0.0406 \\
Steps
& 0.1722 & 0.1337 & 0.0423
& 0.2542 & 0.1207 & 0.0427 \\
\bottomrule
\end{tabular}
}
\caption{\rev{Factor-level summary of aggregate scores pooled across the datasets for which each factor is varied, for two LLaDA-family checkpoints. For each factor, we report the mean, standard deviation (Std), and $\mathrm{Std}/\sqrt{n}$ (SE) over the pooled dataset--setting cells. Because cells come from different datasets, Std mixes between-dataset differences with within-factor sensitivity, and SE is a descriptive summary rather than a sampling interval.}}
\label{tab:factor-summary-backbone}
\end{table}

\begin{table*}[tp]
\centering
\small
\revcolor
\resizebox{\textwidth}{!}{%
\begin{tabular}{lccccccc}
\toprule
Dataset & FVA [95\% CI] & $\eta^2$ & $\omega^2$ & Equal-factor & Equal-level & LOFO & $\rho$ \\
\midrule
\multicolumn{8}{c}{\textbf{LLaDA}} \\
\midrule
PIQA & 0.563\,[0.421,0.672] & 0.244 & 0.052 & 0.266 & 0.509 & 0.493--0.605 & 0.90 \\
WinoGrande & 0.538\,[0.337,0.684] & 0.226 & 0.031 & 0.262 & 0.476 & 0.358--0.566 & 0.90 \\
ARC-Challenge & 0.619\,[0.421,0.753] & 0.289 & 0.106 & 0.374 & 0.529 & 0.479--0.782 & 0.80 \\
HumanEval & 0.788\,[0.734,0.823] & 0.503 & 0.351 & 0.569 & 0.476 & 0.772--0.795 & 1.00 \\
MBPP & 0.773\,[0.744,0.795] & 0.481 & 0.324 & 0.563 & 0.431 & 0.714--0.823 & 1.00 \\
\midrule
\multicolumn{8}{c}{\textbf{LLaDA-1.5}} \\
\midrule
PIQA & 0.583\,[0.437,0.677] & 0.259 & 0.070 & 0.291 & 0.499 & 0.196--0.615 & 0.97 \\
WinoGrande & 0.635\,[0.465,0.740] & 0.304 & 0.124 & 0.369 & 0.418 & 0.626--0.654 & 0.80 \\
ARC-Challenge & 0.765\,[0.596,0.846] & 0.449 & 0.301 & 0.516 & 0.682 & 0.643--0.792 & 0.05 \\
HumanEval & 0.664\,[0.589,0.722] & 0.373 & 0.174 & 0.439 & 0.420 & 0.475--0.756 & 1.00 \\
MBPP & 0.653\,[0.589,0.698] & 0.361 & 0.159 & 0.438 & 0.500 & 0.581--0.721 & 0.80 \\
\bottomrule
\end{tabular}
}
\caption{\rev{FVA uncertainty, observed-grid effect sizes, and design sensitivity for each model--dataset grid. FVA is the setting-weighted contrast of \Cref{tab:fva-by-dataset-backbone} with its 95\% unique-item bootstrap interval; $\eta^2$ and $\omega^2$ are observed-grid effect sizes. ``Equal-factor'' weights factors equally; ``Equal-level'' is the median FVA over random subsamples with the same number of settings per factor; ``LOFO'' is the range of FVA when one factor is left out; $\rho$ is the median Spearman correlation between the full-grid and subsampled factor rankings by aggregate-score range.}}
\label{tab:fva-design-sensitivity}
\end{table*}

\section{\rev{Additional Matched-Item and Choice-Score Diagnostics}}
\label{app:diagnostics}

\rev{Beyond $D_{\mathrm{corr}}$ and $D_{\mathrm{out}}$, \Cref{tab:extra-diagnostics} reports the fraction of configuration pairs whose predictions are both incorrect and differ, the normalized entropy of QA labels across configurations, Fleiss' $\kappa$ for correctness and predictions, and an item-aware setting partial $\eta^2$. For nominal QA labels, Fleiss' $\kappa$ avoids the arbitrary numerical encoding that an intraclass correlation would require. Because factors and settings are deliberately chosen finite-grid conditions, we do not treat them as sampled random effects or report population variance components. Semantic, code-functional, and mixed-effects comparisons are not available and are not claimed.}

\rev{\paragraph{Where the measures diverge.} These diagnostics broadly support, but do not duplicate, the aggregation gap. (i) On ARC-Challenge, whose items have three to five options, full-output disagreement exceeds correctness disagreement (LLaDA-1.5 under MC: $D_{\mathrm{out}}=0.2232$ versus $D_{\mathrm{corr}}=0.1452$, with $D_{\mathrm{joint}}=0.0780$ and $H_{\mathrm{norm}}=0.2104$), so predictions change beyond correctness flips; on the two-option PIQA and WinoGrande, the two coincide by construction. (ii) High agreement can coexist with non-trivial disagreement: the same grid has $\kappa_{\mathrm{corr}}=0.7095$ and $\kappa_{\mathrm{pred}}=0.7005$. (iii) For code, exact-string agreement is much lower than correctness agreement (LLaDA on HumanEval under CFG: $\kappa_{\mathrm{pred}}=0.067$ versus $\kappa_{\mathrm{corr}}=0.525$), reflecting textual rather than functional variation. (iv) The item-aware partial $\eta^2$ can be small (e.g., $0.0067$ for the ARC grid above) even when $D_{\mathrm{corr}}$ is not, because it measures a different quantity. (v) Across the ten model--dataset cases, the Spearman correlation between factor rankings by aggregate-score range and by $D_{\mathrm{out}}$ spans $0.60$--$1.00$: broad correspondence, not equivalence.}

\rev{\paragraph{Choice-score calibration proxies.} \Cref{tab:choice-calibration} summarizes calibration proxies computed from the saved option scores. For LLaDA on WinoGrande under CFG, accuracy varies by only $0.0166$, whereas ECE and Brier score vary by $0.1521$ and $0.1258$; the ECE range exceeds the accuracy range in 22 of 30 grids. These are choice-score proxies rather than token-level calibration, and they answer a narrower question than item-level prediction changes.}

\begin{table*}[p]
\centering
\small
\revcolor
\begin{tabular}{llccccc}
\toprule
Dataset & Factor & $D_{\mathrm{joint}}$ & $H_{\mathrm{norm}}$ & $\kappa_{\mathrm{corr}}$ & $\kappa_{\mathrm{pred}}$ & Partial $\eta^2$ \\
\midrule
\multicolumn{7}{c}{\textbf{LLaDA}} \\
\midrule
\multirow{5}{*}{PIQA} & Precision & 0.0000 & 0.0130 & 0.9758 & 0.9811 & 0.0001 \\
 & Batch Size of MC & 0.0000 & 0.2124 & 0.6787 & 0.7517 & 0.0037 \\
 & CFG & 0.0000 & 0.1108 & 0.8343 & 0.8686 & 0.0014 \\
 & \# of MC & 0.0000 & 0.3769 & 0.4931 & 0.5814 & 0.0152 \\
 & GPUs & 0.0000 & 0.0049 & 0.9875 & 0.9902 & 0.0005 \\
\addlinespace[2pt]
\multirow{5}{*}{WinoGrande} & Precision & 0.0000 & 0.0138 & 0.9744 & 0.9800 & 0.0004 \\
 & Batch Size of MC & 0.0000 & 0.2601 & 0.6132 & 0.6942 & 0.0029 \\
 & CFG & 0.0000 & 0.1609 & 0.7610 & 0.8089 & 0.0011 \\
 & \# of MC & 0.0000 & 0.4602 & 0.3839 & 0.4852 & 0.0161 \\
 & GPUs & 0.0000 & 0.0063 & 0.9838 & 0.9874 & 0.0000 \\
\addlinespace[2pt]
\multirow{5}{*}{ARC-Challenge} & Precision & 0.0102 & 0.0161 & 0.9735 & 0.9685 & 0.0008 \\
 & Batch Size of MC & 0.0523 & 0.1115 & 0.8450 & 0.8262 & 0.0018 \\
 & CFG & 0.0474 & 0.1102 & 0.8360 & 0.8268 & 0.0013 \\
 & \# of MC & 0.0782 & 0.1984 & 0.7283 & 0.7154 & 0.0082 \\
 & GPUs & 0.0034 & 0.0064 & 0.9808 & 0.9827 & 0.0007 \\
\addlinespace[2pt]
\multirow{4}{*}{HumanEval} & Precision & 0.3598 & -- & 0.8577 & 0.5163 & 0.0012 \\
 & CFG & 0.6220 & -- & 0.5251 & 0.0675 & 0.0202 \\
 & GPUs & 0.2561 & -- & 0.9313 & 0.6629 & 0.0305 \\
 & Steps & 0.7348 & -- & 0.3483 & 0.0766 & 0.1867 \\
\addlinespace[2pt]
\multirow{4}{*}{MBPP} & Precision & 0.2793 & -- & 0.8473 & 0.5200 & 0.0019 \\
 & CFG & 0.5940 & -- & 0.4383 & 0.0118 & 0.0524 \\
 & GPUs & 0.1880 & -- & 0.9041 & 0.6614 & 0.0001 \\
 & Steps & 0.6730 & -- & 0.2927 & 0.0607 & 0.2156 \\
\midrule
\multicolumn{7}{c}{\textbf{LLaDA-1.5}} \\
\midrule
\multirow{5}{*}{PIQA} & Precision & 0.0000 & 0.0085 & 0.9836 & 0.9877 & 0.0001 \\
 & Batch Size of MC & 0.0000 & 0.2106 & 0.6774 & 0.7543 & 0.0054 \\
 & CFG & 0.0000 & 0.1167 & 0.8197 & 0.8622 & 0.0006 \\
 & \# of MC & 0.0000 & 0.3894 & 0.4682 & 0.5678 & 0.0165 \\
 & GPUs & 0.0000 & 0.0082 & 0.9783 & 0.9837 & 0.0000 \\
\addlinespace[2pt]
\multirow{5}{*}{WinoGrande} & Precision & 0.0000 & 0.0261 & 0.9530 & 0.9621 & 0.0009 \\
 & Batch Size of MC & 0.0000 & 0.2753 & 0.6042 & 0.6773 & 0.0022 \\
 & CFG & 0.0000 & 0.1602 & 0.7745 & 0.8116 & 0.0035 \\
 & \# of MC & 0.0000 & 0.4644 & 0.3945 & 0.4805 & 0.0103 \\
 & GPUs & 0.0000 & 0.0253 & 0.9367 & 0.9494 & 0.0001 \\
\addlinespace[2pt]
\multirow{5}{*}{ARC-Challenge} & Precision & 0.0068 & 0.0113 & 0.9806 & 0.9779 & 0.0013 \\
 & Batch Size of MC & 0.0462 & 0.1207 & 0.8163 & 0.8152 & 0.0010 \\
 & CFG & 0.0393 & 0.0946 & 0.8592 & 0.8529 & 0.0028 \\
 & \# of MC & 0.0780 & 0.2104 & 0.7095 & 0.7005 & 0.0067 \\
 & GPUs & 0.0009 & 0.0029 & 0.9897 & 0.9920 & 0.0023 \\
\addlinespace[2pt]
\multirow{4}{*}{HumanEval} & Precision & 0.2886 & -- & 0.8152 & 0.4776 & 0.0086 \\
 & CFG & 0.6171 & -- & 0.4861 & 0.0796 & 0.1250 \\
 & GPUs & 0.3171 & -- & 0.7925 & 0.4488 & 0.0137 \\
 & Steps & 0.6209 & -- & 0.3607 & 0.0998 & 0.1699 \\
\addlinespace[2pt]
\multirow{4}{*}{MBPP} & Precision & 0.2213 & -- & 0.8959 & 0.6568 & 0.0031 \\
 & CFG & 0.5768 & -- & 0.5162 & 0.0366 & 0.0645 \\
 & GPUs & 0.2580 & -- & 0.8997 & 0.6054 & 0.0213 \\
 & Steps & 0.5927 & -- & 0.4875 & 0.1114 & 0.1379 \\
\bottomrule
\end{tabular}
\caption{\rev{Additional matched-item diagnostics for every model--dataset--factor grid. $D_{\mathrm{joint}}$ is the fraction of configuration pairs whose predictions are both incorrect and differ; $H_{\mathrm{norm}}$ is the normalized entropy of QA labels across configurations; $\kappa_{\mathrm{corr}}$ and $\kappa_{\mathrm{pred}}$ are Fleiss' $\kappa$ for correctness and predictions; and partial $\eta^2$ is the item-aware setting effect size. For code, $D_{\mathrm{joint}}$ and $\kappa_{\mathrm{pred}}$ compare exact filtered candidate strings and do not measure functional or semantic agreement, and $H_{\mathrm{norm}}$ is not defined.}}
\label{tab:extra-diagnostics}
\end{table*}

\begin{table*}[tp]
\centering
\small
\revcolor
\begin{tabular}{llcccc}
\toprule
Dataset & Factor & Accuracy range & ECE range & Brier range & Confidence range \\
\midrule
\multicolumn{6}{c}{\textbf{LLaDA}} \\
\midrule
\multirow{5}{*}{PIQA} & Precision & 0.0011 & 0.0010 & 0.0001 & 0.0002 \\
 & Batch Size of MC & 0.0370 & 0.0402 & 0.0738 & 0.0052 \\
 & CFG & 0.0158 & 0.0420 & 0.0612 & 0.0371 \\
 & \# of MC & 0.1055 & 0.1179 & 0.2155 & 0.0157 \\
 & GPUs & 0.0016 & 0.0012 & 0.0006 & 0.0004 \\
\addlinespace[2pt]
\multirow{5}{*}{WinoGrande} & Precision & 0.0024 & 0.0050 & 0.0005 & 0.0004 \\
 & Batch Size of MC & 0.0387 & 0.0488 & 0.0600 & 0.0131 \\
 & CFG & 0.0166 & 0.1521 & 0.1258 & 0.1394 \\
 & \# of MC & 0.1200 & 0.1825 & 0.2354 & 0.0681 \\
 & GPUs & 0.0000 & 0.0019 & 0.0003 & 0.0002 \\
\addlinespace[2pt]
\multirow{5}{*}{ARC-Challenge} & Precision & 0.0043 & 0.0041 & 0.0072 & 0.0003 \\
 & Batch Size of MC & 0.0213 & 0.0229 & 0.0239 & 0.0077 \\
 & CFG & 0.0188 & 0.1370 & 0.1315 & 0.1336 \\
 & \# of MC & 0.0563 & 0.0730 & 0.0997 & 0.0235 \\
 & GPUs & 0.0026 & 0.0022 & 0.0025 & 0.0008 \\
\midrule
\multicolumn{6}{c}{\textbf{LLaDA-1.5}} \\
\midrule
\multirow{5}{*}{PIQA} & Precision & 0.0011 & 0.0004 & 0.0006 & 0.0003 \\
 & Batch Size of MC & 0.0446 & 0.0506 & 0.0819 & 0.0071 \\
 & CFG & 0.0114 & 0.0396 & 0.0430 & 0.0377 \\
 & \# of MC & 0.1148 & 0.1273 & 0.2382 & 0.0132 \\
 & GPUs & 0.0005 & 0.0003 & 0.0002 & 0.0004 \\
\addlinespace[2pt]
\multirow{5}{*}{WinoGrande} & Precision & 0.0055 & 0.0056 & 0.0023 & 0.0011 \\
 & Batch Size of MC & 0.0347 & 0.0436 & 0.0578 & 0.0102 \\
 & CFG & 0.0300 & 0.1521 & 0.1467 & 0.1173 \\
 & \# of MC & 0.0947 & 0.1578 & 0.2242 & 0.0568 \\
 & GPUs & 0.0016 & 0.0004 & 0.0007 & 0.0012 \\
\addlinespace[2pt]
\multirow{5}{*}{ARC-Challenge} & Precision & 0.0043 & 0.0045 & 0.0019 & 0.0006 \\
 & Batch Size of MC & 0.0162 & 0.0151 & 0.0146 & 0.0083 \\
 & CFG & 0.0222 & 0.0759 & 0.0697 & 0.0979 \\
 & \# of MC & 0.0461 & 0.0685 & 0.1261 & 0.0224 \\
 & GPUs & 0.0034 & 0.0049 & 0.0033 & 0.0015 \\
\bottomrule
\end{tabular}
\caption{\rev{Choice-score calibration proxies on QA. For each grid, we report the range across configurations of accuracy, expected calibration error (ECE), multiclass Brier score, and mean choice confidence. Scores are a softmax over the saved option log-likelihood estimates; they are not token-logit calibration, and token logits and denoising trajectories were not retained.}}
\label{tab:choice-calibration}
\end{table*}

\section{\rev{Fixed-Protocol, Factorial, and Remasking Experiments}}
\label{app:new-experiments}

\rev{\paragraph{Protocol.} All experiments in \Cref{subsec:fixed-protocol,subsec:interaction} use LLaDA-1.5 at checkpoint revision \texttt{84346fd}\footnote{\raggedright \rev{Full revision hash: \texttt{84346fd91ba60252d260}\allowbreak\texttt{022201ad6fc5a3468fb2}.}}, BF16 model weights, and an FP32 confidence softmax, and run as four processes on four A100 GPUs. For HumanEval, the model generates a 512-token continuation that starts fully masked and is decoded as 16 consecutive 32-token blocks with a linear transfer schedule; temperature is zero, so candidate tokens are argmax tokens. Low-confidence remasking commits the positions with the highest selected-token probabilities, whereas random remasking ranks positions by Torch-uniform scores. PyTorch deterministic algorithms are enabled. For each run, we record the seeds and random streams, dtypes, world size and evaluation order, backend flags, checkpoint and code revisions, prompts, answer extraction, and the exact launch command. In the fixed-protocol repetitions, PIQA uses CFG $0.5$, an MC budget of $128$, and batch size $8$; HumanEval uses CFG $0$, $512$ steps, low-confidence remasking, and batch size $1$. Identical launches use evaluation seed $1234$, and seed-only launches use seeds $2001$--$2010$. Code correctness is the harness's whole-task pass flag; per-test execution vectors were not recorded, so exact-string disagreement is not a measure of functional disagreement.}

\begin{table*}[tp]
\centering
\small
\revcolor
\begin{tabular}{llcccc}
\toprule
Task & Mode & Score & Score SD [95\% CI] & $D_{\mathrm{corr}}$ [95\% CI] & $D_{\mathrm{out}}$ [95\% CI] \\
\midrule
PIQA & Identical & 0.7486 & 0.0000\,[0.0000,0.0000] & 0.0000\,[0.0000,0.0000] & 0.0000\,[0.0000,0.0000] \\
PIQA & Seed-only & 0.7503 & 0.0082\,[0.0067,0.0131] & 0.1212\,[0.1129,0.1305] & 0.1212\,[0.1129,0.1305] \\
HumanEval & Identical & 0.4146 & 0.0000\,[0.0000,0.0000] & 0.0000\,[0.0000,0.0000] & 0.0000\,[0.0000,0.0000] \\
HumanEval & Seed-only & 0.4146 & 0.0000\,[0.0000,0.0000] & 0.0000\,[0.0000,0.0000] & 0.0000\,[0.0000,0.0000] \\
\bottomrule
\end{tabular}
\caption{\rev{Fixed-protocol repetitions of LLaDA-1.5 (ten launches per mode). Score is mean accuracy (PIQA) or pass@1 (HumanEval); brackets give 95\% unique-item bootstrap intervals conditional on the ten runs. For HumanEval, $D_{\mathrm{out}}$ compares exact filtered candidate strings.}}
\label{tab:fixed-protocol}
\end{table*}

\begin{table*}[tp]
\centering
\small
\revcolor
\resizebox{0.6\textwidth}{!}{%
\begin{tabular}{lccc}
\toprule
 & Steps $=128$ & Steps $=256$ & Steps $=512$ \\
\midrule
CFG $=0$ & 0.1890 ($-$0.0264) & 0.3110 ($-$0.0183) & 0.4146 (0.0447) \\
CFG $=1$ & 0.1037 ($-$0.0142) & 0.2561 (0.0244) & 0.2622 ($-$0.0102) \\
CFG $=2$ & 0.0610 (0.0407) & 0.1280 ($-$0.0061) & 0.1402 ($-$0.0346) \\
\bottomrule
\end{tabular}
}
\caption{\rev{HumanEval pass@1 on the $3\times3$ CFG $\times$ Steps grid for LLaDA-1.5; parentheses give the additive interaction residual of each cell.}}
\label{tab:cfg-steps-grid}
\end{table*}

\begin{table*}[tp]
\centering
\small
\revcolor
\resizebox{\textwidth}{!}{%
\begin{tabular}{lccccc}
\toprule
Steps & Low-confidence & Random & $\Delta$ & $D_{\mathrm{corr}}$ & $D_{\mathrm{out}}$ \\
\midrule
128 & 0.1890\,[0.1280,0.2439] & 0.0732\,[0.0366,0.1099] & $-$0.1159\,[$-$0.1707,$-$0.0610] & 0.1402\,[0.0915,0.1951] & 0.9695\,[0.9451,0.9939] \\
512 & 0.4146\,[0.3354,0.4878] & 0.2073\,[0.1524,0.2683] & $-$0.2073\,[$-$0.2744,$-$0.1341] & 0.2805\,[0.2134,0.3476] & 0.9573\,[0.9267,0.9878] \\
\bottomrule
\end{tabular}
}
\caption{\rev{HumanEval pass@1 of LLaDA-1.5 under low-confidence and random remasking. $\Delta$ is random minus low-confidence; $D_{\mathrm{corr}}$ and $D_{\mathrm{out}}$ compare the two policies item by item, with $D_{\mathrm{out}}$ based on exact filtered candidate strings. Brackets give 95\% unique-item bootstrap intervals.}}
\label{tab:remasking}
\end{table*}

\section{\rev{Reporting Checklist and Decision Rule}}
\label{app:checklist}

\rev{\paragraph{Checklist.} We recommend that evaluations of DLMs, and of other models with configurable inference, (i) separate fixed-protocol repetitions, system/execution changes, and deliberate decoding sweeps, and label each experiment accordingly; (ii) prespecify the factors, settings, reference configuration, setting weights, and seeds; (iii) report aggregate scores with their uncertainty together with matched-item correctness and full-output disagreement; (iv) test interactions that are plausible for the comparison being made; and (v) release item-level outputs and configurations where permitted.}

\rev{\paragraph{No universal FVA threshold.} FVA is a relative contrast: it can be high even when the absolute variation is negligible. A universal FVA cutoff for acceptable instability would therefore be misleading.}

\rev{\paragraph{Two-stage decision rule.} \emph{Stage~1.} Before evaluation, fix a tolerance $\tau$ for an absolute measure, such as $D_{\mathrm{corr}}$ or the aggregate-score spread. On the fixed grid, an upper 95\% bound below $\tau$ supports stability, a lower bound above $\tau$ supports material instability, and an interval spanning $\tau$ is inconclusive. Instability whose uncertainty is comparable to a claimed improvement, or that could change a model ranking, should be flagged. \emph{Stage~2.} Only when Stage~1 finds material instability should FVA and per-factor measures be used to localize it; FVA does not measure absolute reliability.}

\rev{\paragraph{Illustration.} With an illustrative tolerance $\tau=0.05$ for $D_{\mathrm{corr}}$ (\Cref{tab:qa-results-backbone,tab:generation-results-backbone}), LLaDA on PIQA is stable under precision changes ($0.0094$ [0.0058, 0.0131]) and shows material instability under CFG changes ($0.0657$ [0.0580, 0.0739]); LLaDA on HumanEval is inconclusive under GPU changes ($0.0305$ [0.0061, 0.0610]). The appropriate tolerance depends on the application and should be fixed before the evaluation.}

\end{document}